\documentclass{article}

\usepackage[dblblindworkshop, final]{neurips_2025}

\usepackage[utf8]{inputenc} % allow utf-8 input
\usepackage[T1]{fontenc}    % use 8-bit T1 fonts
\usepackage{hyperref}       % hyperlinks
\usepackage{url}            % simple URL typesetting
\usepackage{booktabs}       % professional-quality tables
\usepackage{amsfonts}       % blackboard math symbols
\usepackage{nicefrac}       % compact symbols for 1/2, etc.
\usepackage{microtype}      % microtypography
\usepackage{xcolor}         % colors

\usepackage{graphicx}
\usepackage{amsmath}
\usepackage{mathtools}
\usepackage{amsthm}
\usepackage[cal=cm,scr=euler]{mathalpha}
\usepackage{subcaption}
\usepackage{multirow}
\usepackage{wrapfig}
\usepackage[capitalize,noabbrev]{cleveref}
\usepackage{xspace}
\usepackage[export]{adjustbox}

\newcommand{\modelname}{\texttt{HealthPredictor}\xspace}

\title{Predicting Public Health Impacts of Electricity Usage}
\workshoptitle{ResponsibleFM}

\author{
    Yejia Liu\thanks{Equal contribution.} \\
    University of California, Riverside\\
    \texttt{yliu807@ucr.edu}\\
    \And
    Zhifeng Wu\footnotemark[1]\\
    University of California, Riverside\\
    \texttt{zwu178@ucr.edu}\\
    \And
    Pengfei Li\\
    Rochester Institute of Technology\\
    \texttt{pengfei.li@rit.edu}\\
    \And
    Shaolei Ren\thanks{Corresponding author.}\\
    University of California, Riverside\\
    \texttt{shaolei@ucr.edu}
}

\begin{document}

\maketitle

\begin{abstract}
The electric power sector is a leading source of air pollutant emissions, impacting the public health of nearly every community. Although regulatory measures have reduced air pollutants, fossil fuels remain a significant component of the energy supply, highlighting the need for more advanced demand-side approaches to reduce the public health impacts. To enable health-informed demand-side management, we introduce \modelname, a domain-specific AI model that provides an end-to-end pipeline linking electricity use to public health outcomes. The model comprises three components: a fuel mix predictor that estimates the contribution of different generation sources, an air quality converter that models pollutant emissions and atmospheric dispersion, and a health impact assessor that translates resulting pollutant changes into monetized health damages. Across multiple regions in the United States, our health-driven optimization framework yields substantially lower prediction errors in terms of public health impacts than fuel mix-driven baselines. A case study on electric vehicle charging schedules illustrates the public health gains enabled by our method and the actionable guidance it can offer for health-informed energy management. Overall, this work shows how AI models can be explicitly designed to enable health-informed energy management for advancing public health and broader societal well-being. Our datasets and code are released at: \url{https://github.com/Ren-Research/Health-Impact-Predictor}.
\end{abstract}

\section{Introduction}

The electric power sector is a leading source of
air pollutant emissions that affect the public health across
nearly every community \cite{EPA_PowerPlant_Health_LeadingSource_Website},
 yet predicting societal health impacts remains challenging due to the complex relationships between electricity usage, emissions, pollutant dispersion, and health outcomes~\cite{li2024integrated,finkelman2021future}. The urgency of understanding these relationships has intensified with the rapid growth of large energy loads. For instance, the rise of artificial intelligence (AI) and large language models has led to unprecedented energy demand from data centers
~\cite{calma2023ai}. This trend, combined with the increasing electrification of transportation and industrial processes, makes electricity usage a critical sector for mitigating public health impacts, a critical topic of social well-being.

\begin{wrapfigure}[18]{r}{.42\textwidth}
     \vspace{-0.1cm}	
    \centering
    \includegraphics[width=1\linewidth]{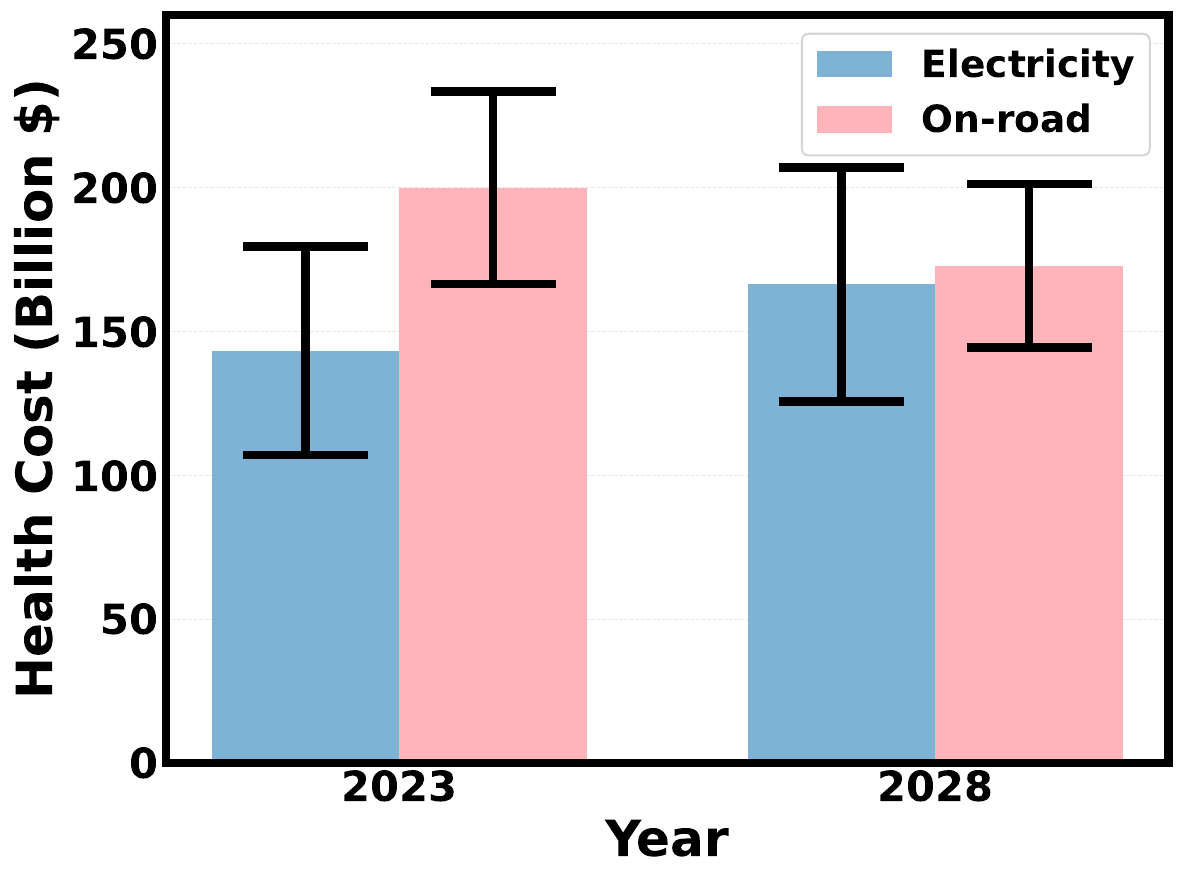}
     \vspace{-0.6cm}	    
     \caption{{Total public health costs of electricity generation and on-road emissions in the contiguous U.S. in 2023 and 2028 \cite{Health_COBRA_EPA_Website}. The error bars represent high and low estimates provided by COBRA using two different exposure-response models.}}
    \label{fig:on_road_comparsion}
\end{wrapfigure}

Electricity consumption directly impacts public health through air pollution from fossil fuel power plants, which remain one of the largest industrial polluters~\cite{EPA_PowerPlant_Health_LeadingSource_Website,WHO2021AirQuality}. Despite strict  regulations reducing power sector emissions, fossil fuel plants continue to be ``a leading source of air, water, and land pollution that affects communities nationwide'' in the United States, as reported by the EPA ~\cite{epa2024cobra}. Analysis using the EPA's COBRA modeling tool \cite{EPA-COBRA} indicates that health costs from electricity are on track to rise, rivaling those of on-road emissions in 2028 as shown in Figure~\ref{fig:on_road_comparsion}.

Coal-fired power plants are among the fossil fuel facilities with the most adverse health effects, with their PM\textsubscript{2.5} emissions estimated to have caused approximately 460,000 excess deaths between 1999 and 2020 \cite{henneman2023mortality}.\footnote{While greenhouse gas emissions such as carbon dioxides may also be broadly classified as global air pollutants, their impacts on public health are often second- or third-order and different from
the immediate health outcomes resulting from criteria air pollutants such
as PM\textsubscript{2.5} \cite{EPA_CleanAirAct_Summary_Website,Health_COBRA_EPA_Mannual,henneman2023mortality}. In this paper, we focus on the public health impacts of criteria air pollutants without including greenhouse gas emissions.}
Despite their significant health impact,
they remain a key component of the U.S. electricity mix. Importantly, the U.S. EIA projects that even by 2050, fossil fuels will still account for a significant share of electricity generation, with coal power generation remaining around 180 billion kWh under the alternative scenario where power plants are allowed to operate subject to rules existing before early 2024 \cite{eia_aeo2025_browser}.
The continued reliance on fossil fuels means that even though the U.S. is among
the leading countries in clean energy development, the associated health risks cannot be overlooked. 

In Europe, the European Environment Agency estimates that air pollution from large industrial facilities, led by power and heat plants, causes health and environmental damages equivalent to about 2\% of EU GDP~\cite{Health_Europe_AirPollution_PowerPlant_1Percent_GDP_2024_Update_2025}. Globally, dependence on coal and other fossil fuels for electricity has remained largely steady over the past forty years  \cite{ElectricityMix_FossilFuel_OurWorldinData_2024}.
Therefore, it is crucial to predict and mitigate these health impacts through targeted interventions from demand-side management alongside supply-side transitioning to cleaner grids.

Importantly, the relationship between electricity use and health impacts offers unique opportunities for intervention because a large portion of  the electricity demand is dynamically \textit{controllable}, unlike other natural pollution sources or weather patterns. This controllability enables proactive demand-side management by tapping into energy load flexibilities,
e.g., scheduling data center workloads or coordinating electric vehicle (EV) charging schedules in residential sectors.

Prior research has evolved from epidemiological pollution-health correlations to advanced machine learning (ML) for modeling complex public health impacts. Some early studies have quantified the health effects of air pollutants, particularly particulate matter, laying the foundation for air dispersion models like COBRA~\cite{EPA-COBRA} and InMAP~\cite{inmap} by leveraging models like Gaussian dispersion equations and chemical transport simulations to estimate pollutant spread and associated health risks. More recent advances  have transformed this research landscape. \cite{seng2021spatiotemporal} demonstrated the effectiveness of LSTM networks in air quality assessment and pollution forecasting, achieving higher accuracy than traditional statistical methods. Researchers have developed foundation models that integrate diverse data sources to forecast comprehensive atmospheric composition~\cite{bodnar2024foundationmodelearth}.

Nonetheless, the existing studies mostly address isolated aspects of the problem, either assessing health impacts from air pollutants or modeling energy-to-emissions conversion~\cite{finkelman2021future, bodnar2024foundationmodelearth}. Given that fossil fuel generation will remain a substantial part of the power grid for the foreseeable future, there is a critical gap to design a new AI model capable of connecting demand-side
usage directly to health outcomes. Such predictions would provide valuable signals to users, enabling them to take informed actions to mitigate air pollution to protect public health by leveraging demand-side flexibilities.

We present a new domain-specific AI model, \modelname, an end-to-end pipeline that quantifies the public health impacts of electricity consumption. Our model integrates three key components: a fuel mix predictor that forecasts the proportional contribution of different energy sources to electricity generation, an air quality converter that models pollutant emissions and their atmospheric dispersion, and a health impact assessor that translates pollution changes into monetary health costs. By combining these components with a health-driven optimization framework, \modelname enables prediction of health impacts from electricity usage to inform demand-side management. We demonstrate the effectiveness of our approach through an EV charging case study, where users can determine optimal charging schedules to minimize adverse health outcomes. 
Our approach bridges the critical gap between electricity consumption and health outcomes, providing actionable insights for both individuals and system operators. In addition, we release the datasets we have collected and processed to help advance the field of research by addressing the limitations of fragmented and dispersed data from various sources as observed in previous works~\cite{bodnar2024foundationmodelearth}.

\section{Related Works}

\textit{Energy system modeling} typically focuses on technological and economic characteristics, often incorporating health damage in an aggregated and simplified manner~\cite{subramanian2018modeling, montenegro2021beyond}. These approaches rarely provide granular insights into the direct health impacts of electricity generation. Some methodologies focus on optimizing energy systems to reduce emissions \cite{Health_Power_SupplySide_Optimization_Francesca_Dominici_Harvard_arXiv_2024_battiloro2024healthbasedpowergridoptimization}, but their health impact assessments tend to remain indirect or high-level, missing the opportunity for detailed, localized health assessments~\cite{nkyekyer2019use,li2024integrated}. 

\textit{Epidemiological studies} have made significant contributions to understanding the relationship between health and air pollution~\cite{epa2024cobra}. For example, \cite{hall2024regional} has conducted a comprehensive regional impact assessment of air quality improvement, while \cite{chen2016effects} developed log-linear models for quantifying asthma hospitalizations based on particulate matter levels, demonstrating the direct correlations between air quality and health outcomes in urban environments. Although these studies provide a valuable foundation for linking environmental pollutants with health impacts, they lack an integrated framework that connects power systems directly to health outcomes. Air pollution dispersion modeling also plays a critical role in supporting these epidemiological studies. The study \cite{pantusheva2022air} systematically reviewed computational fluid dynamics approaches for urban air pollution modeling, and  has developed InMap \cite{inmap}, a specialized model for analyzing air pollution interventions, accounting for complex atmospheric chemical interactions. More advanced dispersion modelings
are also available \cite{bodnar2024foundationmodelearth}. 
While these models offer insights into the dispersion of pollutants, they typically do not link air pollutants to users' energy decisions, and thus do not provide actionable insights for individuals to make improvements.

Recent advances have integrated machine learning with environmental health research. ~\cite{seng2021spatiotemporal} explored the use of LSTM network for air quality assessment and pollution forecasting, demonstrating the potential of data-driven approaches. Additionally, the emerging field of \textit{health-informed computing}, exemplified by works like~\cite{han2024unpaidtollquantifyingpublic}, seeks to quantify the broader societal impacts of technological systems, providing a methodological foundation for the future research on the health consequences of electricity generation, mainly with respect to the advancement of AI and the development of large data centers.

The existing studies do not consider end-to-end frameworks that directly quantify the health impacts of electricity consumption across residential or industrial sectors, with a few notable exceptions \cite{Health_Benefit_kWh_EPA_Website,WattTime_HealthDamage_Intro_Website}.
Specifically, the EPA reports annual average health damages associated with electricity consumption for various U.S. regions, which can help inform energy efficiency programs or spatial planning for renewable deployment \cite{Health_Benefit_kWh_EPA_Website}. However, the absence of temporal variation limits its usefulness for dynamic demand-side energy management. In contrast, \cite{WattTime_HealthDamage_Intro_Website} provides real-time health impact signals for electricity use, but these signals reflect only marginal damages (that is, the incremental health impact from consuming an additional unit of electricity), and the underlying methodology is proprietary, leaving limited room for external verification or scientific scrutiny. These limitations point to the need for more comprehensive and transparent approaches that cover diverse fuel sources and regions while providing actionable guidance for system operators and individual users.

\section{Background and Problem Formulation}

In this section, we review the background and introduce formulations related to health impact assessments from the use of power generation fuel mix.

\subsection{Air Pollutants for Health Impact}
Ambient/outdoor air pollution is now recognized as the second largest risk factor for noncommunicable diseases, contributing to approximately 4.2 million premature deaths globally each year \cite{Health_AirPollution_Ambient_4dot2_million_PrematureDeath_2019_WHO_Website}. 
Thus, air quality is a critical determinant of human health, shaped by the presence of specific gases and particulate matter in the atmosphere. Six pollutants are recognized as primary contributors to air quality degradation: carbon monoxide (CO), nitrogen oxide (NO), nitrogen dioxide (NO\textsubscript{2}), sulfur dioxide (SO\textsubscript{2}), ozone (O\textsubscript{3}), and particulate matter in three size categories, PM\textsubscript{1}, PM\textsubscript{2.5}, and PM\textsubscript{10}~\cite{WHO2021AirQuality}. 
These pollutants originate from various sources, including fossil fuel combustion and industrial processes~\cite{bodnar2024foundationmodelearth}. The long-distance transport of these pollutants amplifies their public health impact, particularly for vulnerable populations such as the elderly and individuals with preexisting conditions. Adverse health outcomes, including premature mortality, asthma exacerbation, and cognitive decline, also result in substantial societal costs through increased hospitalizations and medication use~\cite{Grande2021CognitiveDecline,EPAHealthEffects}.

\subsection{Air Dispersion}
\label{s:formulation-dispersion}
Establishing meaningful relationships between emission sources and their health impacts is non-trivial. Typically, the first step is to determine the spatial and temporal distribution of pollutants in the area. This process usually involves mathematical models with varying spatial resolutions that solve the governing dispersion-advection equations. By integrating emission data with meteorological inputs, dispersion models can estimate pollutant concentrations at specific receptor points~\cite{pantusheva2022air,bodnar2024foundationmodelearth}.

Assume there are \( K \) types of air pollutants and \( M \) receptor regions of interest. Let \( \mathscr{P}_s = (\mathscr{P}_{s, 1}, \dots, \mathscr{P}_{s, K}) \) denote the quantities of \( K \) types of air pollutants at the emission source. For receptor \( i \), the corresponding quantities are represented by \( \mathscr{P}_r^i = (\mathscr{P}^{i, 1}_{r}, \dots, \mathscr{P}^{i, K}_r) \). A general dispersion model can be formulated as:
\begin{equation}
\mathscr{P}_{r}^1, \dots, \mathscr{P}_{r}^M = D_{\boldsymbol{w}}(\mathscr{P}_s),
\label{eq:dispersion}
\end{equation}
which gives the amount of \( K \) types of air pollutants at receptor region \( i = 1, \dots, M \), i.e., \( \mathscr{P}_r^i = (\mathscr{P}^{i, 1}_{r}, \dots, \mathscr{P}^{i, K}_r) \). The parameter \( \boldsymbol{w} \) captures factors such as geographical conditions, characteristics of emission source, and meteorological data~\cite{han2024unpaidtollquantifyingpublic, tessum2017inmap}.

Despite rapid advancements in mathematical models, the uncertainty still exists due to the complex interplay between emission sources and meteorological conditions~\cite{pantusheva2022air, cordova2021air}. While emission models require detailed anthropogenic data, meteorological predictions depend on both measurements and simulations to capture atmospheric turbulence.

\subsection{Measuring Health Impacts}
\label{s:formulation-healthImp}
The relationship between changes in adverse health effects and changes in air pollution exposure can be quantified using epidemiological studies~\cite{epa2024cobra}. For example, the rate of asthma hospitalizations can be modeled as a log-linear function of particulate matter levels~\cite{chen2016effects}. Specifically, for a receptor region \( i \), the change in the number of adverse health effects \( \Delta Y^i \) can be expressed as:
\begin{equation}
\Delta Y^i = Y_0^i \times \text{POP}^i \times \left( 1 - e^{-\alpha^T \Delta \mathscr{P}_r^i} \right),
\label{eq:log-linear}
\end{equation}
where \( Y_0^i \) is the baseline incidence rate for the health outcome at receptor \( i \), \( \text{POP}^i \) is the population exposed at the receptor $i$, \( \alpha \) is the concentration-response coefficient derived from epidemiological studies, and \( \Delta \mathscr{P}_r^i = (\Delta \mathscr{P}_r^{i, 1}, \dots, \Delta \mathscr{P}_r^{i, K}) \) is the change in pollutant concentrations at receptor \( i \).

\subsection{Converting Health Impacts into Monetary Valuation}
Health Impact Assessment (HIA) often requires converting health outcomes into monetary values to enable cost-benefit analysis and facilitate policy decision-making~\cite{trejo2019quantifying, epa2024cobra}. We denote these values by \( v^i = (v^{i,1}, \dots, v^{i, H}) \), where \( H \) represents the number of different types of health impacts (e.g., premature mortality, asthma attacks) at receptor \( i \). Commonly used methodologies for this conversion include estimating the economic value of a statistical life (VSL), as proposed by the Organization for Economic Cooperation and Development (OECD)~\cite{oecd2016economic}, and quality-adjusted life years (QALYs)~\cite{hall2024regional,lomas2016pharmacoeconomic}.
It is important to note that the health impacts associated with pollutant exposure at time 
$t$ reflect effects that unfold over subsequent years, typically within a five-year window.

{Our work} establishes the connection between electricity generation at a source \( s \) over time steps \( t = 1, \dots, T \), denoted as \( E^s_t = (E^{s,1}_1, \dots, E^{s,F}_T) \), where \( F \) represents the number of different fuel mix sources (such as oil and gas), and the resulting economic health outcomes \( v^i_t \) at receptor \( i \) at time step \( t \).

\section{Methods}\label{methods}
\begin{figure*}[t]
    \centering    \includegraphics[width=0.9\linewidth]{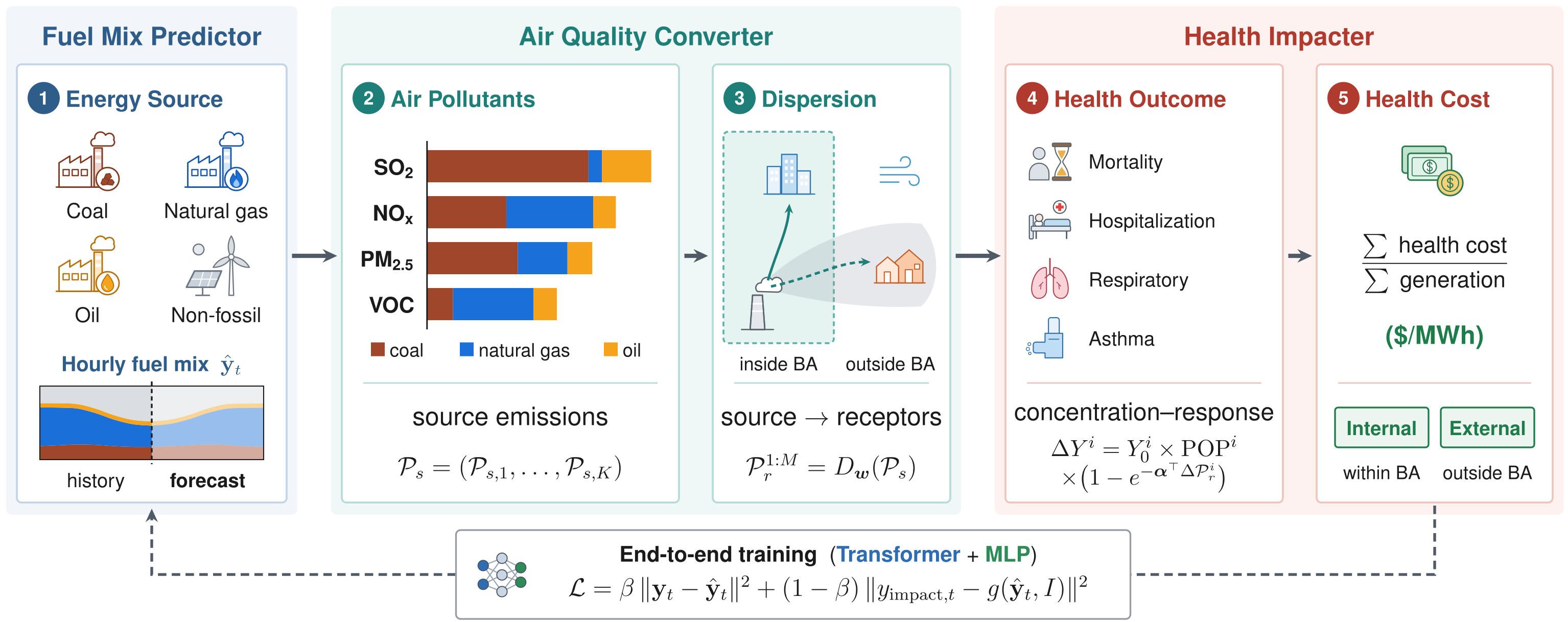}
    \caption{Overview of our end-to-end \modelname. The pipeline begins with energy contribution, $E'_t$, from various sources (e.g. gas, coal). It then models pollutant dispersion (e.g. SO\textsubscript{2}, PM\textsubscript{2.5}) to receptors. Finally, it quantifies the resulting health impacts by monetary cost metrics (\$/MWh).
    }
    \label{fig:pipeline}
\end{figure*}

Our methodology integrates diverse datasets and modeling approaches into an  end-to-end pipeline, which links electricity consumption to public health outcomes based on the power generation fuel mix pattern, named the {\modelname}. As shown in Figure~\ref{fig:pipeline}, the framework consists of three core modules: the Fuel Mix Predictor, the Air Quality Converter, and the Health Impacter.

\subsection{Modeling Framework}
\subsubsection{Fuel Mix Predictor}

A fuel mix predictor is the starting point of our pipeline, as it directly influences the health impacts of electricity generation. Variations in the fuel mix, driven by factors such as demand, renewable availability, and regulations, cause fluctuations in pollutant emissions, which in turn impact human health outcomes like respiratory illnesses and premature deaths~\cite{bodnar2024foundationmodelearth}. Therefore, accurately predicting the fuel mix is crucial to estimate these emissions and assess their potential health impacts.

The goal of the fuel mix predictor is to estimate the future fuel mix (e.g., coal, oil, gas) utilized for electricity generation across the next time horizon based on historical data on the grid's fuel mix. The fuel mix data is inherently time-variant~\cite{mosavi2019state}. Several machine learning approaches have been proposed to model these chaotic and nonlinear time-series relationships, including the LSTM networks to capture temporal dependencies in energy forecasting~\cite{karasu2022crude}. Hybrid approaches, such as combining wavelet basis functions (WBF), sparse autoencoders (SAE), and LSTM, also aim to improve prediction accuracy by integrating multiple advanced models~\cite{qiao2021combination}. For our predictor, we opt for the Transformer-based architecture in favor of its superior ability to capture long-range dependencies.

\subsubsection{Air Quality Converter} 
The Air Quality Converter is tasked with transforming predicted fuel mix data into quantifiable air pollutant emissions in our pipeline.

\paragraph{Pollutants Estimation} 

Each fuel type has distinct emission factors that determine the amount of pollutants emitted per unit of energy generated. These factors can vary depending on combustion technology, fuel quality, and operating conditions. Additionally, regional differences, such as local fuel types, regulatory standards, and emission control technologies, can further influence these emission factors~\cite{epa2024cobra}. By obtaining the fuel mix predictions from the fuel mix predictor, we can estimate pollutant emissions by multiplying the fuel mix with the corresponding emission factor for each fuel.

\paragraph{Dispersion Modelling}  
Modelling the dispersion of air pollutants is a critical step in understanding the relationship between emissions and their resulting concentrations in the atmosphere. This process provides insight into how pollutants spread, dilute, and interact with environmental conditions, ultimately determining their impact on air quality and public health. Two primary types of pollutants are usually considered in dispersion modeling. One is the primary pollutants, such as directly emitted particulate matter (e.g., PM2.5). It usually exhibits a linear relationship with source emissions. These pollutants can be effectively modeled using tools like AERMOD, a steady-state Gaussian plume dispersion model recommended by the EPA~\cite{epa2024cobra}, which calculates pollutant concentrations by accounting for environmental variables such as wind speed, atmospheric stability, and source characteristics. The other is the secondary pollutants,  such as O\textsubscript{3} formed from precursors NO and NO\textsubscript{2}, involve more complex, non-linear relationships with emissions. Their formation results from chemical reactions in the atmosphere, influenced by factors like sunlight and temperature. To model these interactions, chemical transport models (CTMs) that simulate atmospheric chemistry and transport
processes are commonly used~\cite{inmap}. 

The general relationship between pollutant concentrations at receptor sites and emissions from sources can be expressed by Eq.~(\ref{eq:dispersion}). In this formulation, \( D_{\boldsymbol{w}}(\mathscr{P}_s) \) encapsulates the complex dispersion process, where \( \boldsymbol{w} \) accounts for factors such as geographical conditions, emission source characteristics, and meteorological data.  In our framework, 
we consider a simplified modeling approach as used in the EPA's COBRA \cite{Health_COBRA_EPA_Mannual}, use the prevailing weather pattern,
and model the dispersion function \( D_{\boldsymbol{w}}(\mathscr{P}_s) \)  using a custom neural network layer designed to approximate the transformation \( f(E_s) \), which represents the transformation applied to the emissions $E_s$. This neural network layer takes as input the pollutant quantities at the emission source (\( \mathscr{P}_s \)) along with relevant environmental features, and predicts pollutant quantities at the receptor regions (\( \mathscr{P}_r^i \) for \( i = 1, \dots, M \)). The neural network parameters, corresponding to \( \boldsymbol{w} \), are trained to minimize the discrepancy between predicted and observed concentrations.  By leveraging the neural network-based dispersion modeling, we aim to provide a flexible framework for predicting pollutant concentrations, while noting
that more advanced models that incorporate real-time weather conditions
are also possible \cite{bodnar2024foundationmodelearth}.

\subsubsection{Health Impacter}  
The health impact component aims to quantify the changes in adverse health effects resulting from variations in air pollution exposure, following the results obtained from the air quality converter. We measure the health impact in dollars per megawatt-hour (\$/MWh). This measurement reflects the economic cost of health impacts associated with electricity generation~\cite{oecd2016economic}. The most widely utilized functional form in criteria air pollutant concentration-response modeling is the log-linear model, as introduced in Eq.~\eqref{eq:log-linear}. This model is well-suited to capture non-linear relationships between pollutant concentrations and health risks. In addition to the log-linear model, linear models are also applied in specific cases, such as when evaluating the health impacts of certain pollutants (e.g., SO\textsubscript{2} or PM\textsubscript{2.5}) or within specific demographic groups where simpler proportional relationships may better describe the data~\cite{epa2024cobra}.

The health impact modeling process aligns closely with established epidemiological frameworks and tools such as COBRA~\cite{epa2024cobra}, which quantify the proportional increase in health risks due to incremental changes in pollutant concentrations. These models incorporate concentration-response functions derived from epidemiological studies, enabling a robust estimation of health risks, such as premature mortality and respiratory illnesses~\cite{epa2024cobra}. By leveraging these functions, our pipeline calculates the economic valuation of health impacts per unit of electricity generation.

\subsection{End-to-End Training}
\paragraph{Loss Function Design}
We design a loss function for a health-informed learning pipeline by incorporating the health impact  directly into the optimization process.

Let \( y_t \) be the true fuel mix at time \( t \) and \( \hat{y}_t \) be the predicted fuel mix at time \( t \) by the fuel mix predictor. 
We denote \( g(\cdot) \) as the function that estimates health impacts measured by \$/MWh based on the fuel mix predictor and other applicable features 
$I$ including pollutant levels and spatial features. 
The loss function is formulated as:

\begin{equation}
\mathcal{L}(\hat{y_t}|y_t,y_{\text{impact}, t}) =  \beta\left\lVert y_t - \hat{y_t}) \right\rVert^2 + (1-\beta) \left\lVert y_{\text{impact}, t} - g(\hat{y}_t, I) \right\rVert^2 ,
\label{eq:loss}
\end{equation}

where $y_\text{impact}$ denotes the true value of the health impact, while $g(\hat{y}_t, I)$ represents the predicted health impact, based on the predicted fuel mix and a series of models that convert the corresponding pollutants into health impacts, and $\beta$ is a hyperparameter to balance the prediction accuracy of the fuel mix and the health impact. It is worth noting that $y_{\text{impact}, t}$ here is general, which can refer to local health impacts, global health impacts, or a combination of both, depending on the context and the extent of the dispersion of pollutants in the atmosphere. This flexibility allows our pipeline to account for both direct localized effects and broader regional or global consequences of air pollution.

In training, our pipeline learns to predict health impacts from fuel mix data through multiple stages, using a customized loss function specifically tailored to incorporate accurate health impact prediction, as shown in Eq.~(\ref{eq:loss}). The input to our pipeline is a sequence of fuel mix data, which represents the distribution of fuel types over time (e.g., hourly fuel mix data for a year). The output is the predicted health impact, expressed as a monetary value per unit of energy (\$/mWh).

\section{Experiments}
\label{s:main-exps}
In this section, we implement our pipeline and develop methods, including health impact-driven approaches, to demonstrate the effectiveness and flexibility of our pipeline. This implementation serves as the foundation for the case study carried out in Section~\ref{s:ev-charging}, which signals EV users to reduce adverse health impacts during charging.

\textbf{Datasets}\quad
Our analysis uses fuel mix data from U.S. Energy Information Administration (EIA)~\cite{EIA-fueldata} with emission rates derived from eGRID~\cite{EPA-eGRID} and health cost estimates from COBRA~\cite{EPA-COBRA} and health impact data (\$/MWh) based on estimates, as detailed in Appendix~\ref{appendix:data}.
Our experiments are conducted on three major power regions: California (CISO), Texas (ERCO), and the Mid-Atlantic (PJM). These regions reflect diverse characteristics in grid operations, emission profiles, and public health impact patterns. The dataset includes six input features, such as fuel mix percentages and time period, and two output features: internal (within-BA) and external (outside-BA) health impacts,
where BA refers to ``balancing authority''. More details and additional empirical results are provided in Appendix~\ref{appendix:data}.

\textbf{Model Construction}\quad
For the fuel-mix predictor, we develop a Transformer-based architecture tailored to fuel mix time-series data. To model the non-linear conversion of emissions to health impacts, we utilize a 3-layer Multi-Layer Perceptron (MLP) . Detailed model architectures are provided in Appendix~\ref{appendix:additional_empirical}.

\textbf{Implementation Details}\quad
Our pipeline predicts three outputs: \textit{Fuel Mix}, \textit{Health Impact (Internal)}, and \textit{Health Impact (External)}. The latter two are derived from fuel mix predictions and account for the dispersion of air pollutants beyond the source region~\cite{epa2024cobra}. The ``Internal" captures the total health cost within a BA’s jurisdiction, while the ``External" reflects the cost in all counties outside that domain. The loss function in Eq.~(\ref{eq:loss}) for the case study can then be rewritten as
\begin{equation}
\mathcal{L}(\hat{y_t}|y_t,y_{\text{impact}, t})= \beta\left\lVert y_t - \hat{y_t} \right\rVert^2 + \frac{1-\beta}{2} \bigl(\left\lVert y_{\text{impact},i,t} - g_i(\hat{y}_t, I) \right\rVert^2 + \left\lVert y_{\text{impact},e, t} - g_e(\hat{y}_t, I) \right\rVert^2 \bigr)
\end{equation}
where $y_{\text{impact},i,t}$ represents the within-region health impact, and $y_{\text{impact},e,t}$ represents the external (outside-region) health impact at time $t$. In our experiments, we observed no clear justification to prioritize internal versus external health impacts for regions. Therefore, to avoid notation clutter, we set both hyperparameter values for within-region and outside-region health impacts to $\frac{1 - \beta}{2}$.

In our experiments, predictions are made across time windows ($T$) of 24 and 72 hours. In addition to our Transformer-based models, we also implement LSTM-based variants of both the Fuel-mix-driven Opt and Health-driven Opt methods as baselines for comparison. We choose LSTM as a baseline due to its proven effectiveness in capturing temporal dependencies in time-series data~\cite{lstm_temporal_prediction}. Further details on data splitting and optimizer hyperparameters are moved to Appendix~\ref{appendix:additional_empirical}.

\begin{figure}
    \centering
    \subfloat[CISO]{
    \includegraphics[height=0.24\linewidth, valign=b]{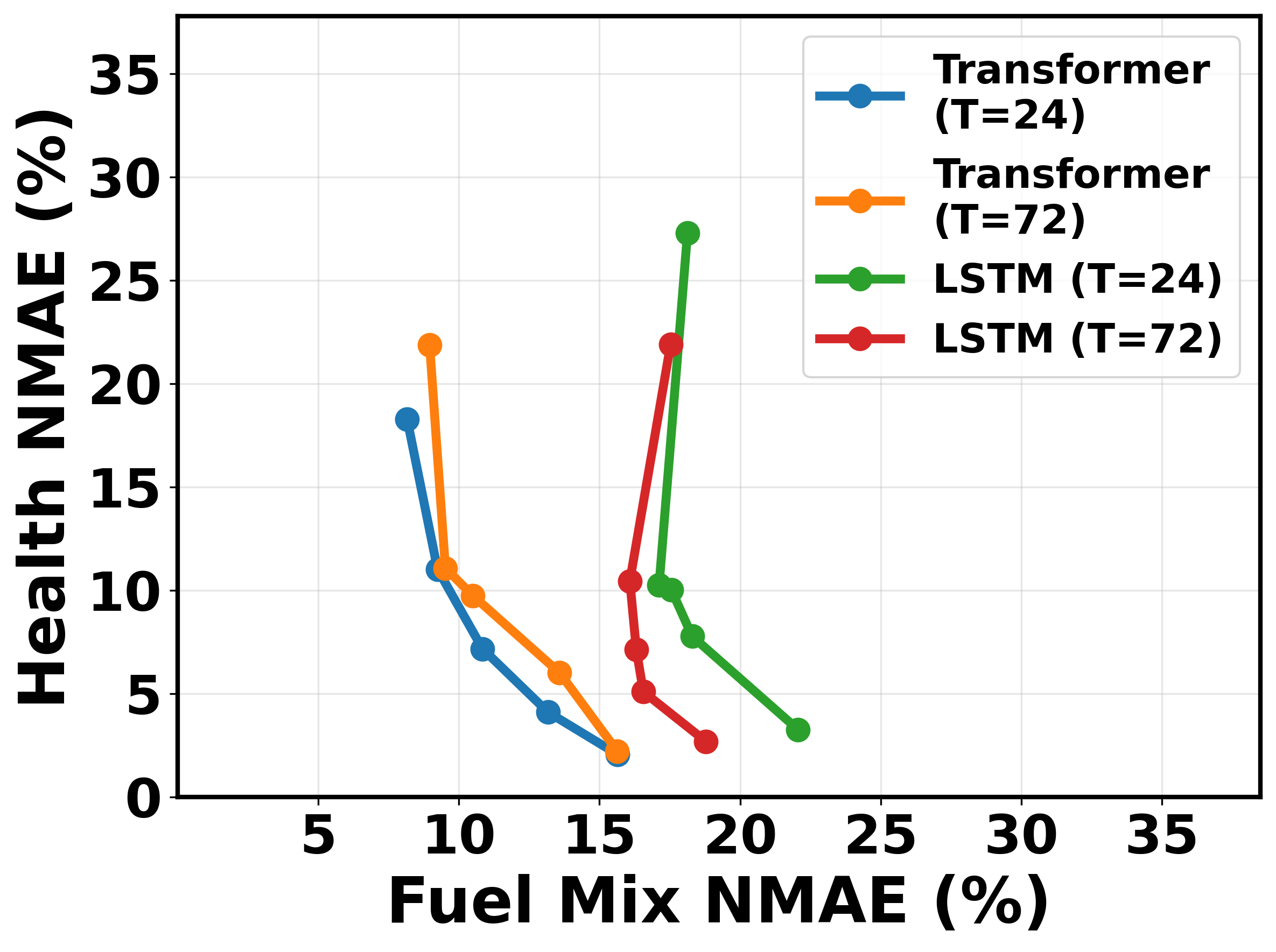}
    }
    \subfloat[ERCO]{
    \includegraphics[height=0.24\linewidth, valign=b]{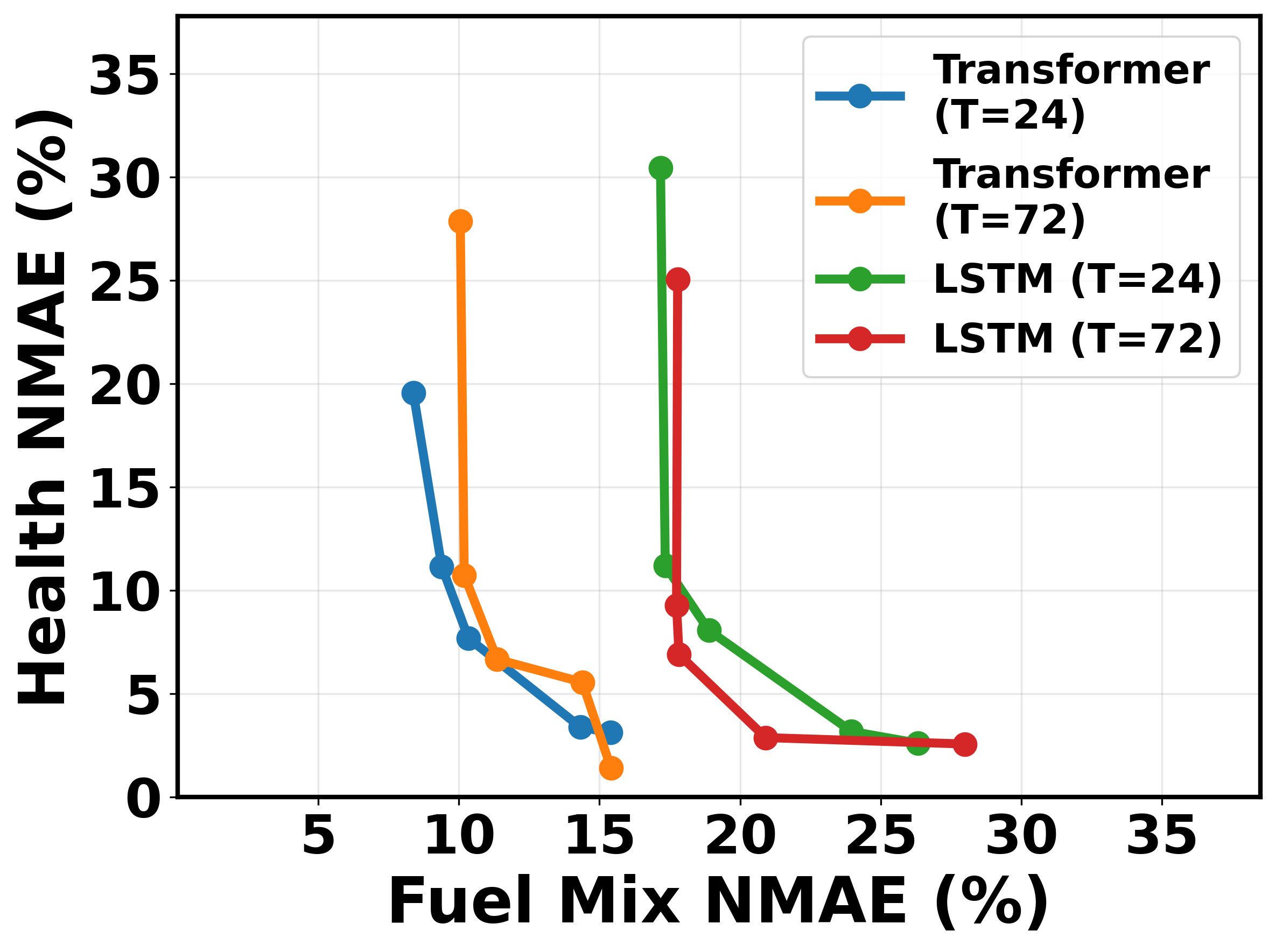}
    }
    \subfloat[PJM]{
    \includegraphics[height=0.24\linewidth, valign=b]{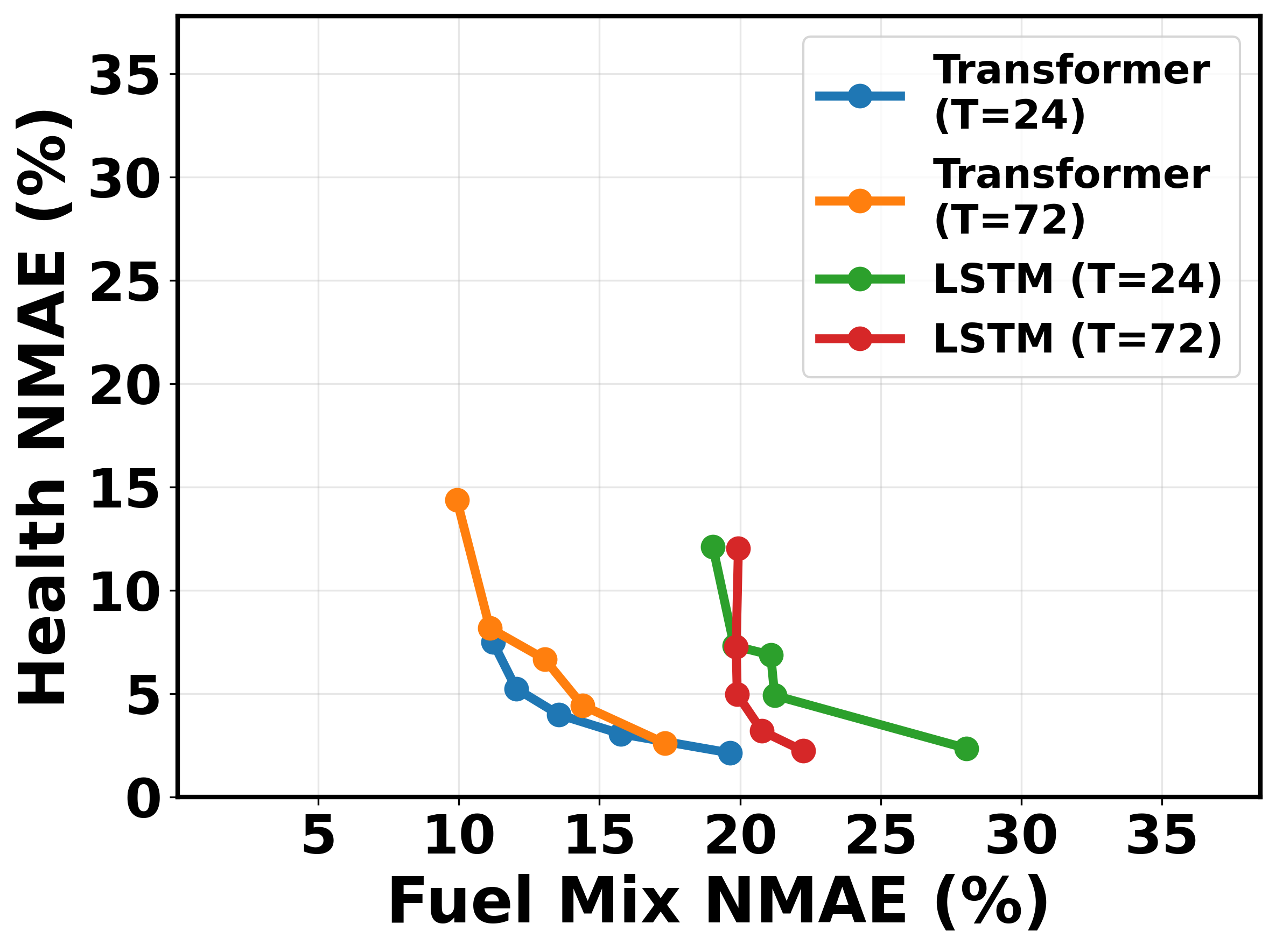}
    }
    \caption{{Trade-off between health impact prediction and fuel mix prediction accuracy across CISO, ERCO, and PJM regions. NMAE refers to Normalized Mean Absolute Error.    
    The top-left points of each curve correspond to the \textit{Fuel-mix-driven Opt}, while the bottom-right points represent the \textit{Health-driven Opt}.}}
    \label{fig:main-res}
    \end{figure}

\textbf{Main Results}\quad
{We evaluate our pipeline by sweeping $\beta$ across (0, 1) in Eq.~\eqref{eq:loss}, where $1-\beta$ controls the weight assigned to health impact optimization (evenly split between internal and external impacts as $\frac{1-\beta}{2}$ each). Values of $\beta$ close to 1 (maximum 0.998 in our case) correspond to \textit{Fuel-mix-driven Opt}, which prioritizes fuel mix prediction accuracy, while smaller values yield \textit{Health-driven Opt}, which prioritizes minimizing health impact prediction error.
We cannot set 
$\beta=1$ because doing so would prevent the model from learning air dispersion or health outcomes. In that case, estimating health impacts would require relying entirely on ground-truth modeling tools, which can be time-consuming to run, especially for complex regulatory-grade models.

Figure~\ref{fig:main-res} illustrates the trade-off between fuel mix and health impact prediction performance across CISO, ERCO, and PJM regions. Note that the reported Health NMAE represents the aggregated error of both internal and external health impacts. First, \textit{Health-driven Opt} consistently achieves lower health impact NMAE. Second, Transformer-based architectures generally outperform LSTM baselines across both optimization objectives and prediction windows (T=24, 72), demonstrating superior modeling capacity for this task.} Importantly, incorporating
the downstream health impact into the predictor is necessary to ensure accurate signaling to users for health-informed energy management.

\section{A Case Study of Health-Aware EV Charging}
\label{s:ev-charging}
While EVs can eliminate tailpipe emissions, the increasing adoption of EVs can still potentially
impact the public health through emissions associated with electricity generation. Scheduling EV charging strategically can play a critical role in reducing harmful emissions, thereby mitigating public health risks and also supporting power system stability~\cite{filote2020environmental}.

\modelname provides a real-time health impact {signal} for EV users over the next few hours based on electricity usage patterns. These predictions evaluate health impacts caused by electricity usage, expressed in units of \$/MWh, guide users to identify optimal charging times, helping to reduce pollutant emissions and their health impacts. By delivering quantifiable and actionable insights, \modelname empowers users to make informed decisions, effectively reducing the health risks associated with electricity usage. In our case study, different charging schedule strategies along with their corresponding numerical results are presented and analyzed.

\textbf{Setups}\quad
For each EV with a total charging demand of $z$, 
the charging occurs within a time frame starting at $t_{I}$ and ending at $t_{E}$. To optimize this process, we discretize the interval $[t_{I}, t_{E}]$ into  time slots and implement a binary charging scheme $B=(b_{t_{I}},\cdots, b_{t_{E}})$. Each element ${b}_{t}$ in $B$ is either $1$, indicating charging at time $t$, or $0$, indicating no charge. The (constant) charging rate is denoted by $c$. Considering the health impact ${h}_t$ at time $t\in[t_{I}, t_{E}]$, the goal to minimize the total charging health impact by determining the optimal charging schedule for EV $j$ can be expressed as follows,
\begin{align}\label{eq:charging-eq-1}
    \begin{split}
        \min_{B=(b_{t_{I}},\cdots, b_{t_{E}})}
          \sum_{t=t_{I}}^{t_{E}} c \cdot b_t \cdot {h}_t,\;\;\;
        s.t. 
        \quad 
       \sum_{t=t_{I}}^{t_{E}} c\cdot b_t=z.
    \end{split}
\end{align}

\textbf{EV-Charging Datasets}\quad
We use the publicly available ACN-Data~\cite{acndata}, which provides real-time charging details (e.g., arrival/departure times, energy delivered), to estimate power demand and charging rates for EVs in residential areas.
To approximate the available residential charging time window, we leverage data from the National Household Travel Survey (NHTS)~\cite{NHTS}. We assume the distributions of the initial charging time  and end time align with the NHTS distributions of home arrival and departure times, respectively. For the health impact predictions ${h}_t$, we use the empirical results from Section~\ref{s:main-exps} on different regions.

\textbf{Simulation Results}\quad
We evaluate several charging strategies: \textit{First Hours}, which charges during the earliest available hours after arriving; \textit{Latest Hours}, which charges during the latest available hours before departure; and \textit{Continuous Charging}, which involves non-interruptible charging continuously from an optimal starting time  to satisfy the demand while minimizing the overall health impact.

\begin{figure*}
\centering
\begin{subfigure}[b]{0.32\linewidth}
    \includegraphics[width=\linewidth]{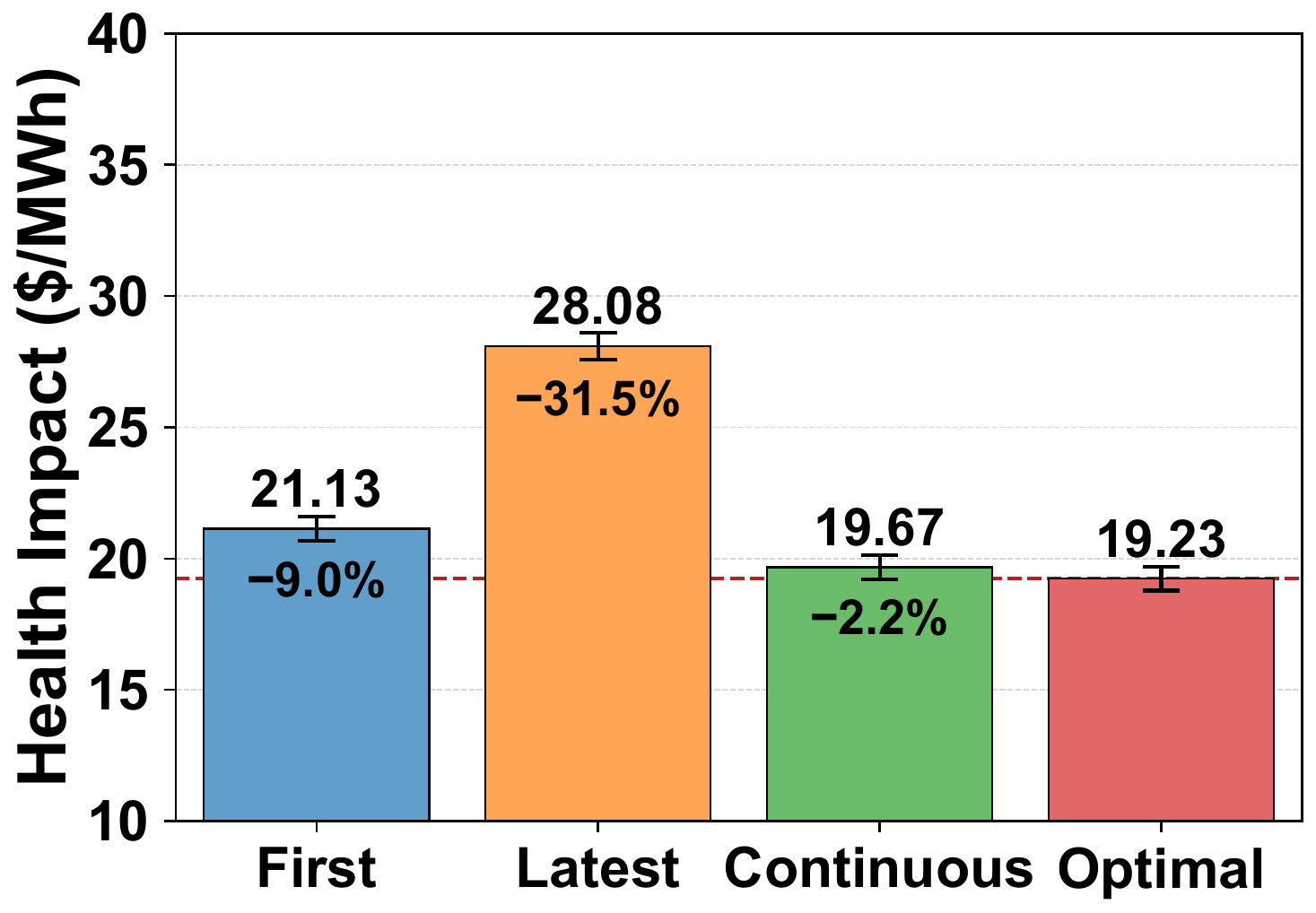}
    \caption{CISO}
  \end{subfigure}
  \begin{subfigure}[b]{0.32\linewidth}
    \includegraphics[width=\linewidth]{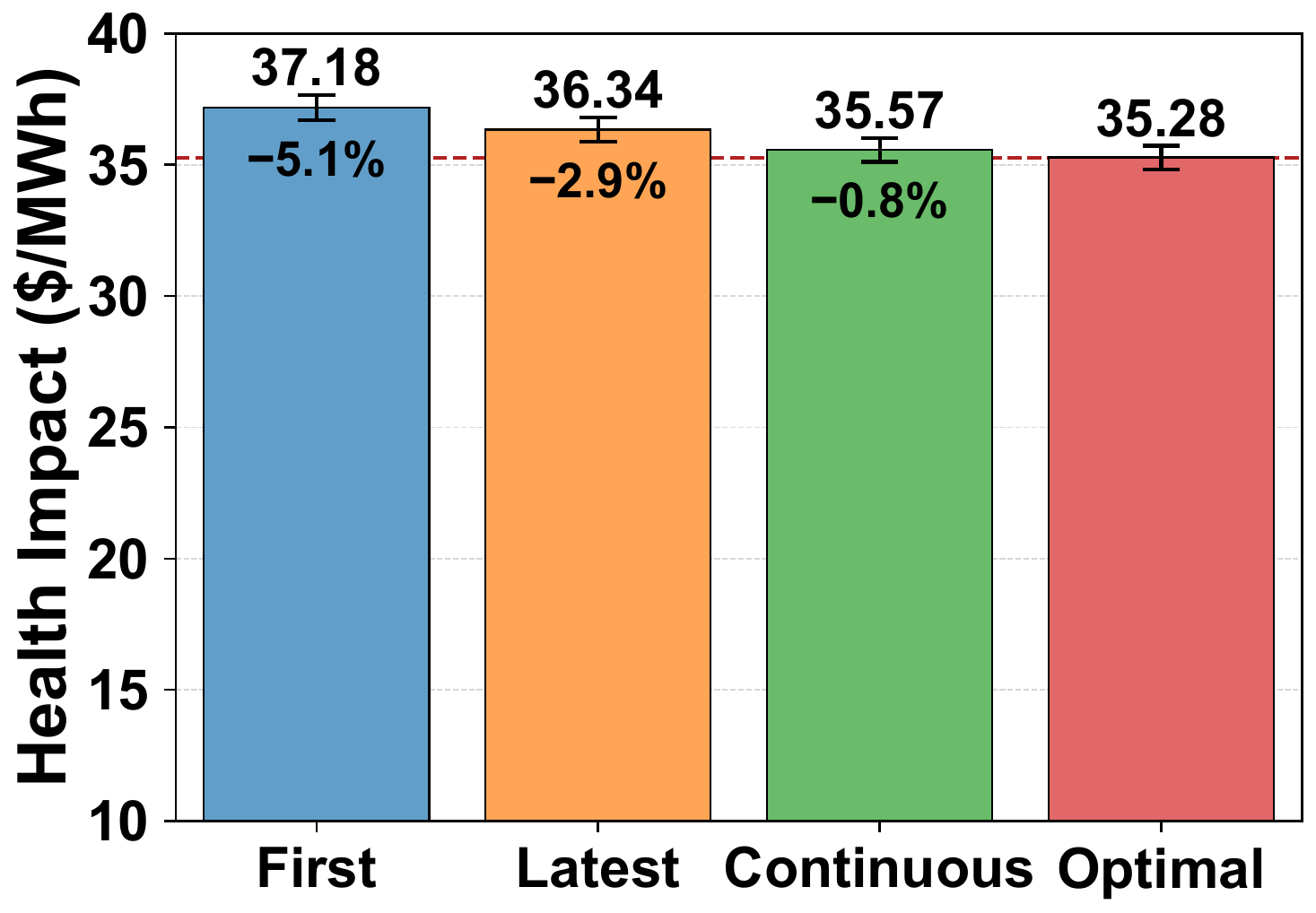}
    \caption{PJM}
  \end{subfigure}
  \begin{subfigure}[b]{0.32\linewidth}
    \includegraphics[width=\linewidth]{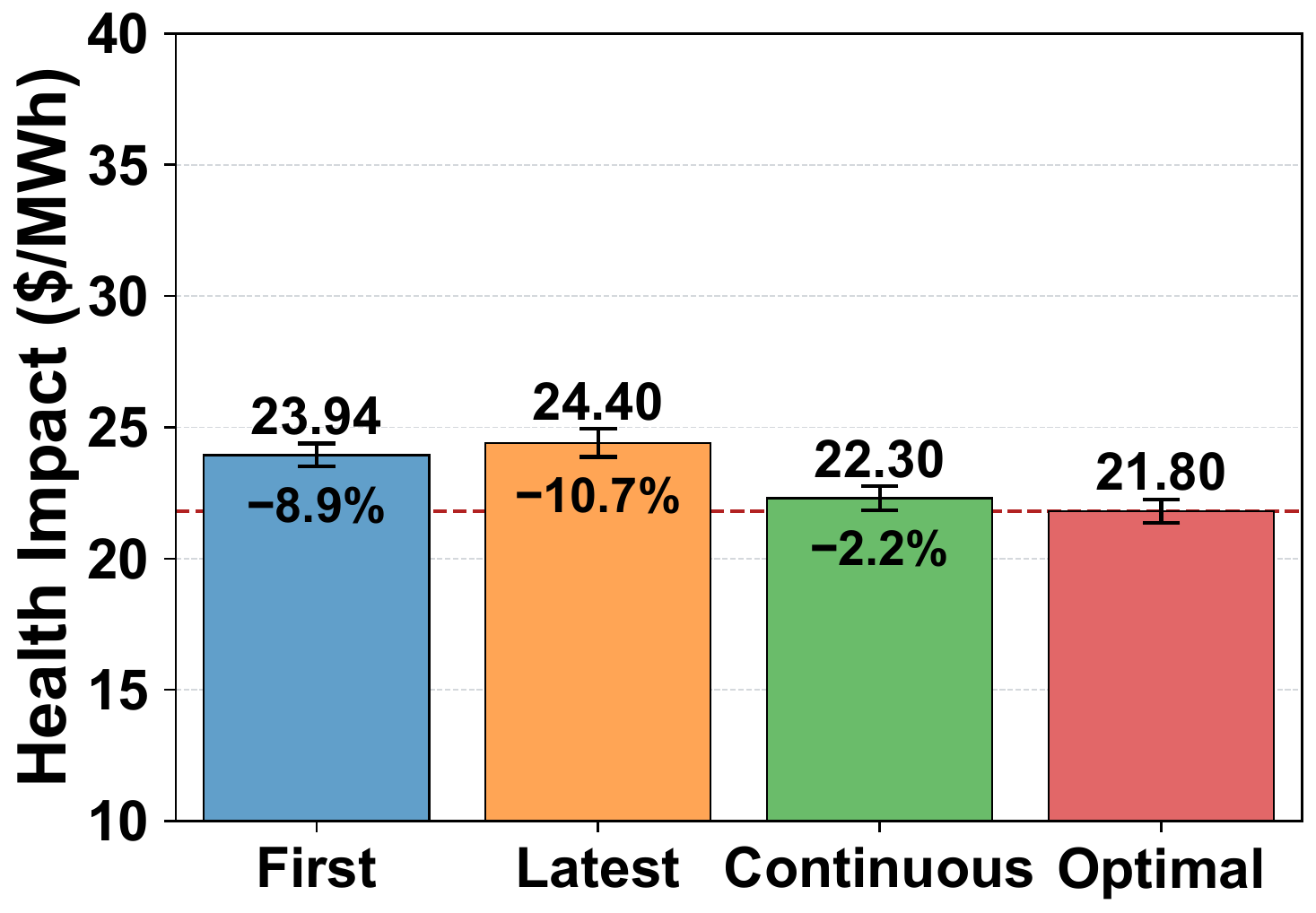}
    \caption{ERCO}
  \end{subfigure}
 \caption{Simulation results of using different EV charging strategies based on health impact predictions in CISO, PJM and ERCO regions. With the provided prediction signals from the \modelname, EV users can choose the \textit{optimal} hours to charge their vehicles, achieving the greatest adverse health outcomes reduction compared to other charging strategies. Each bar shows the mean daily health impact over the 365 days of 2023, and its error bar shows the 95\% confidence interval of the mean. The percentage inside each bar is the reduction achieved by \textit{Optimal} relative to that strategy, and the dashed line marks the \textit{Optimal} level.}
 \label{fig:ev-re-main}
\end{figure*}

In the simulation, we use predictions of both internal and external health impacts of Health-driven Opt method from Section~\ref{s:main-exps} to calculate ${h}_t$.
Figure~\ref{fig:ev-re-main} compares the total health impacts generated throughout the entire charging process. By optimizing the charging schedule using Eq.~(\ref{eq:charging-eq-1}) which selects optimal charging hours based on health impact predictions, great reductions in total health impacts can be achieved. Specifically, across the CISO, PJM, and ERCO regions, our approach reduces total health impacts by $\sim$3–32\% compared to the \textit{First Hours} and \textit{Latest Hours} strategies, and by $\sim$1–2\% compared to \textit{Continuous Charging}.

\section{Conclusion}\label{main:conclusion}
This work introduces a novel approach to bridging the gap between electricity consumption decisions and their public health implications. Our \modelname demonstrates that incorporating health impact considerations into electricity usage predictions can lead to substantial reductions in adverse health outcomes. The effectiveness of our approach is validated across three U.S. regions and through a practical case study on EV charging optimization, showing potential health impact reductions of 1-32\% compared to other charging strategies. By providing quantifiable health impact predictions, \modelname enables more health-informed decision-making for both individuals and system operators. 

\textbf{Limitations.} We acknowledge several limitations in our study. For example, our predictions only consider relatively short time windows and do not extend to long-term scenarios. Additionally, while we use the EPA's air dispersion model as the ground truth, there may still be high level of uncertainty in air dispersion due to the complex interplay between emission sources and meteorological conditions~\cite{pantusheva2022air, cordova2021air}.

\section*{Acknowledgement}
This work was supported in part by the U.S. NSF under
the grant
 CCF-2324941.

\bibliographystyle{plain}

\appendix 
\section{Appendix}\label{appendix}
\subsection{Datasets Collection and Preparation Details}\label{appendix:data}
We here document details on constructing the comprehensive datasets that link the fuel mix usage of power generation with health outcomes for training the \modelname.

Beyond the end-to-end training design, it is important to highlight that the datasets required to train our pipeline are not only complex but also fragmented and labor-intensive to construct. These data come from heterogeneous sources with inconsistent schemas and geographic granularity. For example, aligning fuel categories across agencies (e.g., 8 types in EIA vs. 40 in EPA eGRID) requires systematic mapping, and spatial integration between balancing authorities (BAs) and county-level health data involves optimized point-in-polygon indexing. We have created datasets that link hourly fuel mix compositions with corresponding health costs, covering all 67 BAs in the U.S. for the latest available year, totaling 586,920 data points. We release these datasets with our code to support future research and practical applications.

\begin{figure}[htbp!]
    \centering
    \subfloat[CISO]{ 
        {\includegraphics[height=0.23\textwidth]{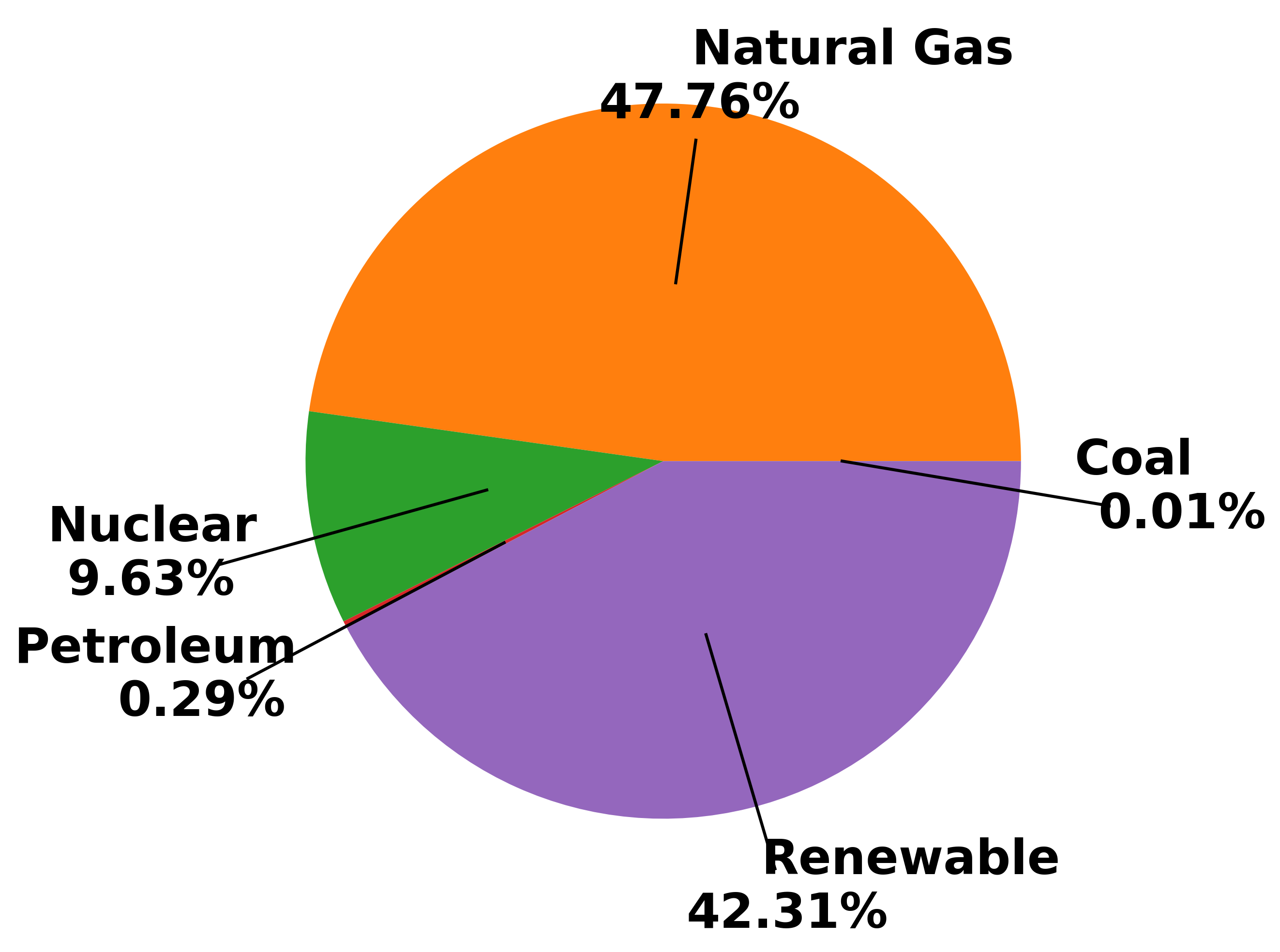}\label{fig:ciso-mix}
        }
    }
    \subfloat[PJM]{
        {\includegraphics[height=0.23\textwidth]{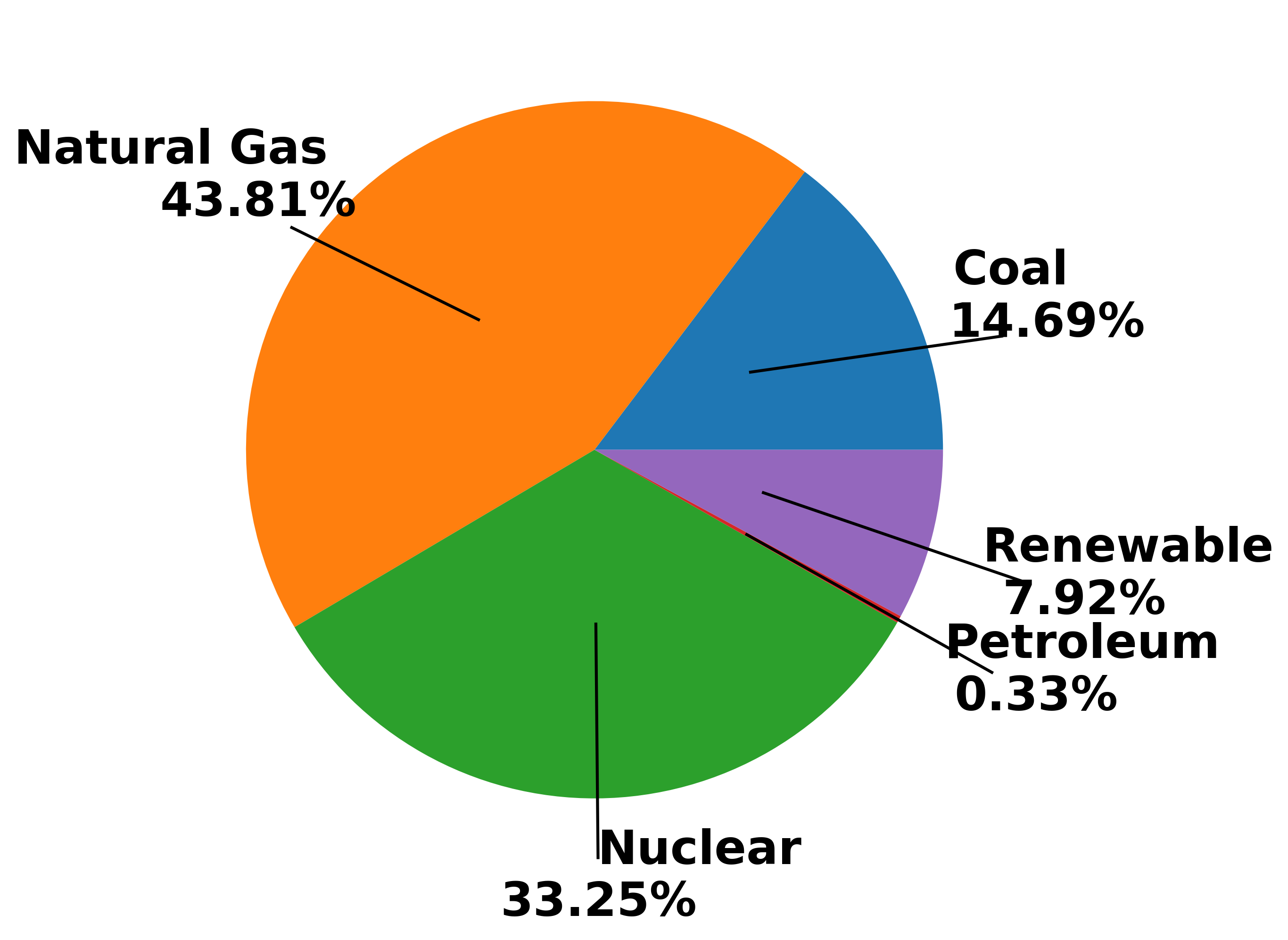}\label{fig:pjm-mix}
        }
    }
    \subfloat[ERCO]{
        {\includegraphics[height=0.23\textwidth]{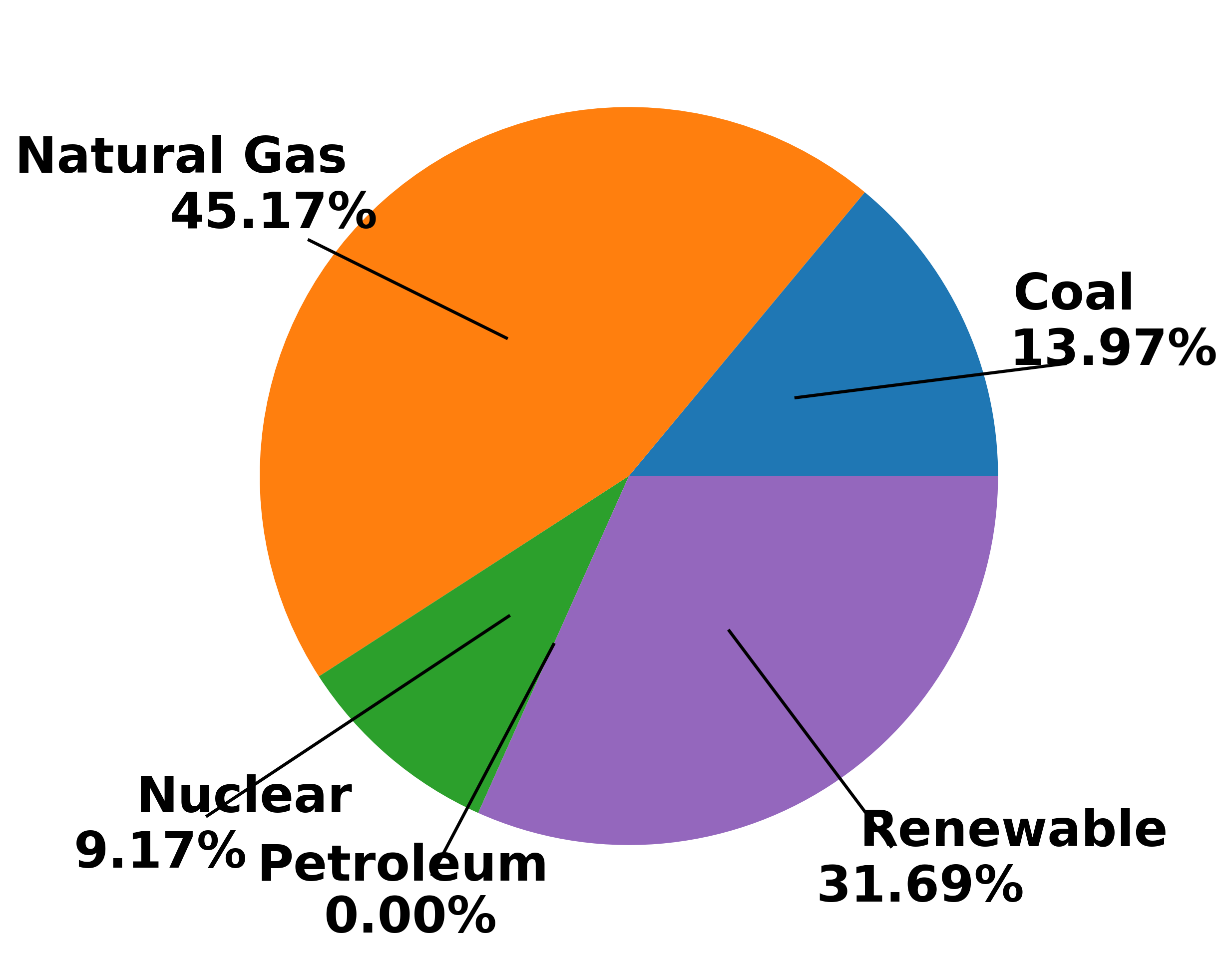}\label{fig:erco-mix}
        }
    }
    \caption{Distribution of the energy generation mix by different fuel types in CISO, PJM, and ERCO.}
    
    \label{fig:region-energy-mix}
\end{figure}

Our primary dataset consists of hourly generation fuel mix data and their corresponding internal and external health costs per megawatt-hour (MWh) for the selected geographical regions in the most recent year. To construct this comprehensive dataset, we employed a systematic approach encompassing data acquisition, processing, and analysis through the following procedures:
\paragraph{Step I: Acquisition of Hourly Generation Mix Data}
We collect the hourly generation fuel mix data of latest years from the U.S. Energy Information Administration (EIA)~\cite{EIA-fueldata}. The U.S. EIA provides electricity generation data organized by balancing authorities (BAs, a functional role defined by the North American Electric Reliability Corporation~\cite{NERC}) rather than state boundaries. This kind of organization is preferred as BAs align more closely with the operational structure of the power grid, providing a more accurate representation of how electricity is generated and managed across regions. In Figure~\ref{fig:region-energy-mix}, we report the distribution of different fuel types for the regions we studied: CISO, PJM, and ERCO, averaged hourly throughout the year.
For ERCO, the petroleum consumption is processed as 0\% in our analysis due to the EIA including it within the broader "Other" category without specific data. According to~\cite{EIA-fueldata}, petroleum usage in power generation is generally minimal in ERCO, so excluding it as a separate category does not affect the overall fuel mix analysis for our methods. In processing the fuel mix data, missing data points are addressed using a two-step imputation approach. The primary method involves interpolation based on adjacent hourly data. When such data are unavailable, missing values are substituted by averaging corresponding time points from the nearest available days, taking advantage of daily cyclical patterns in the fuel mix.

\begin{figure}[htbp!]
    \centering
    \subfloat[CISO]{ 
        {\includegraphics[height=0.23\textwidth]{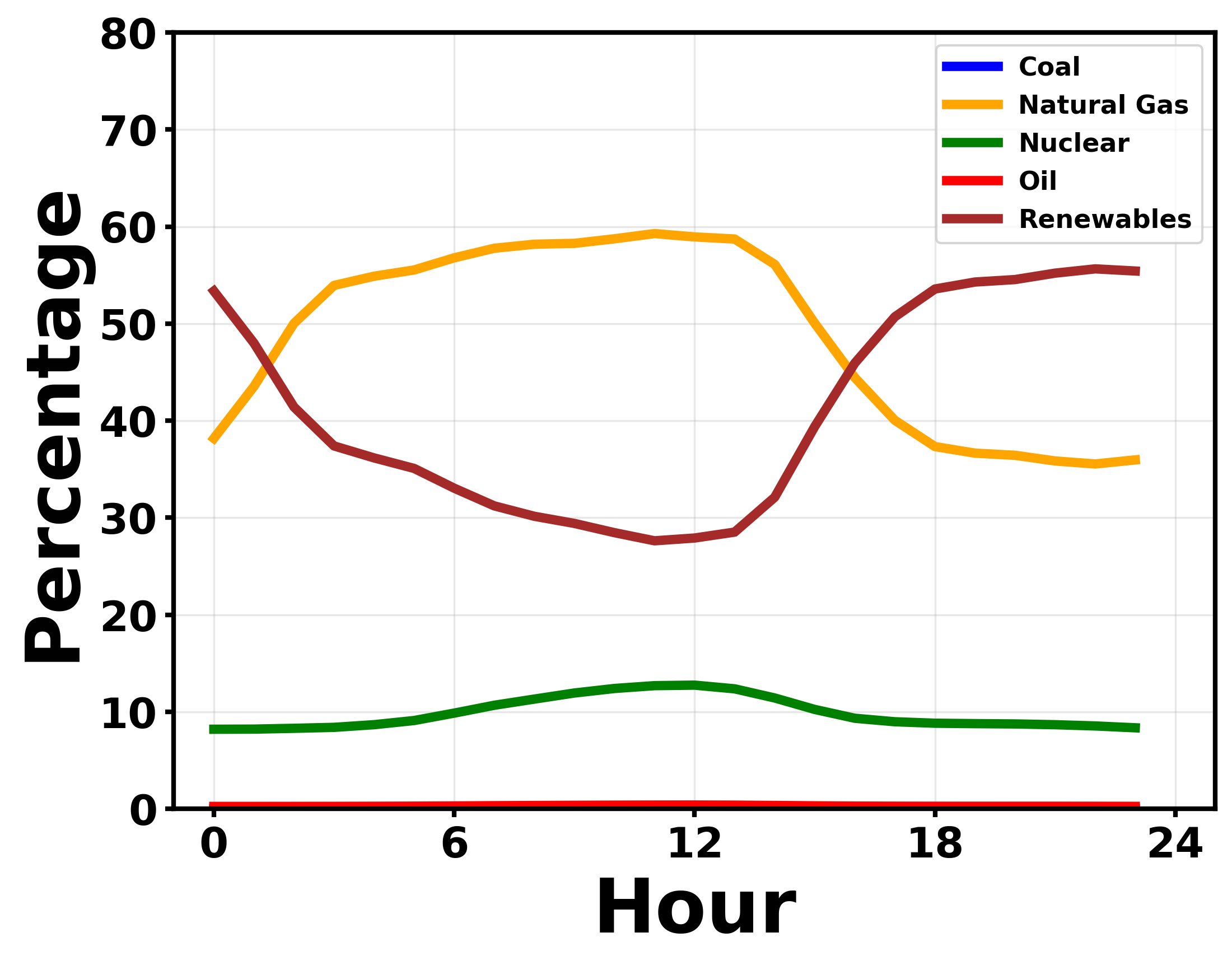}\label{fig:ciso-mix-hourly}
        }
    }
    \subfloat[PJM]{
        {\includegraphics[height=0.23\textwidth]{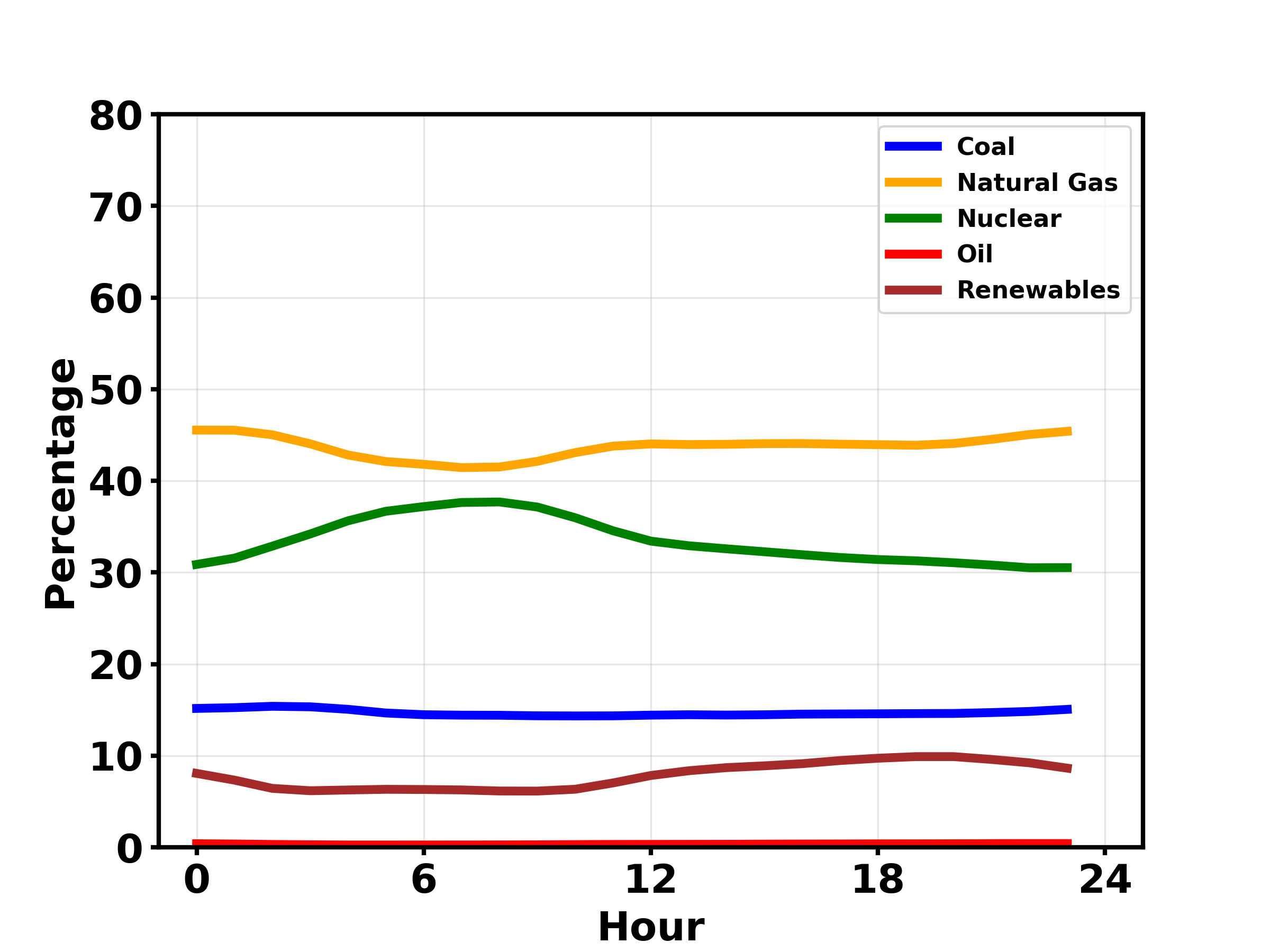}\label{fig:pjm-mix-hourly}
        }
    }
    \subfloat[ERCO]{
        {\includegraphics[height=0.23\textwidth]{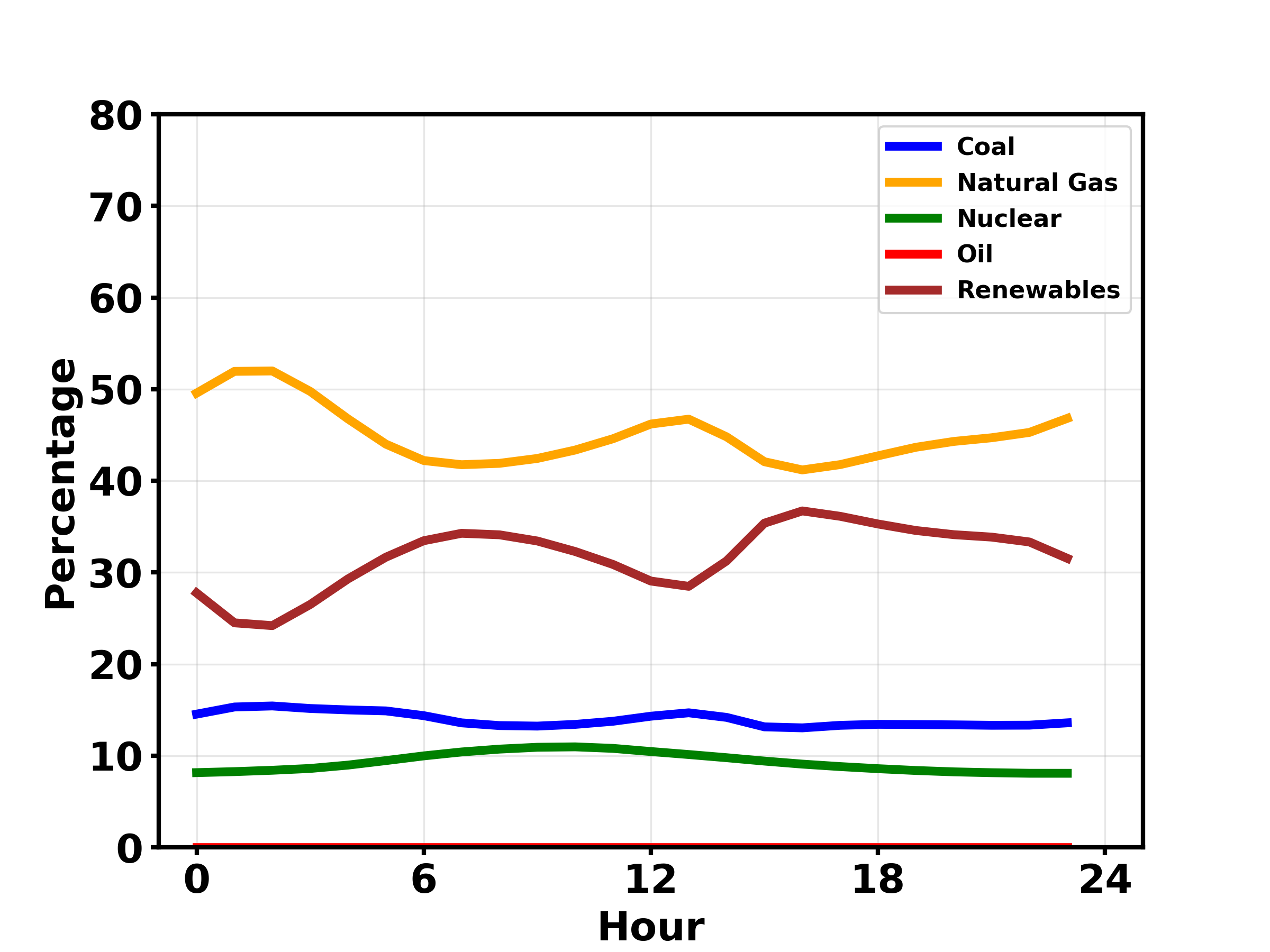}\label{fig:erco-mix-hourly}
        }
    }
    \caption{Percentage of hourly energy generation by fuel type in CISO, PJM, and ERCO.}
    
    \label{fig:region-energy-mix-hourly}
\end{figure}

\begin{figure}[htbp!]
    \centering
    \subfloat[CISO]{ 
        {\includegraphics[height=0.21\textwidth]{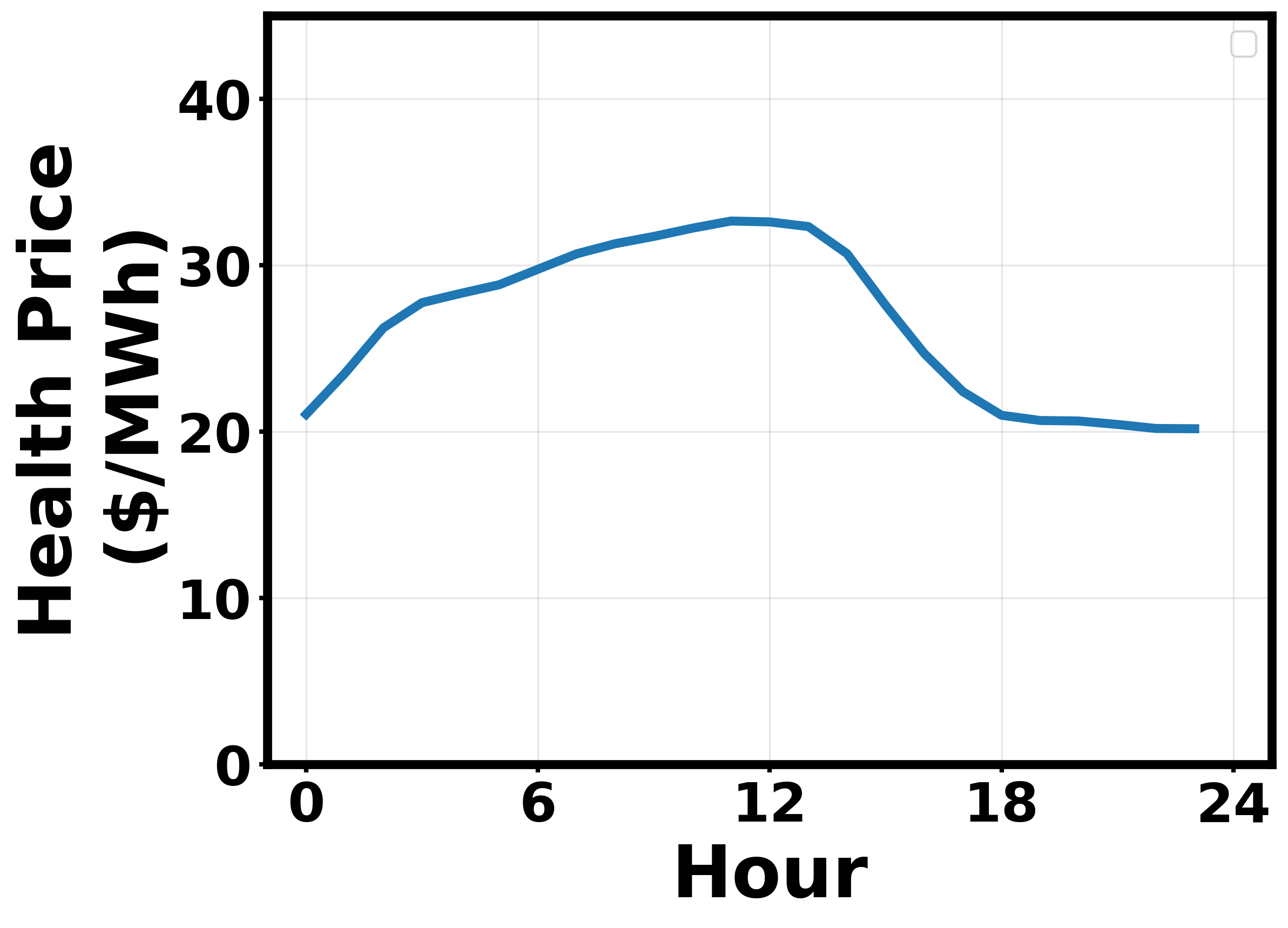}\label{fig:ciso-health-hourly}
        }
    }
    \subfloat[PJM]{
        {\includegraphics[height=0.21\textwidth]{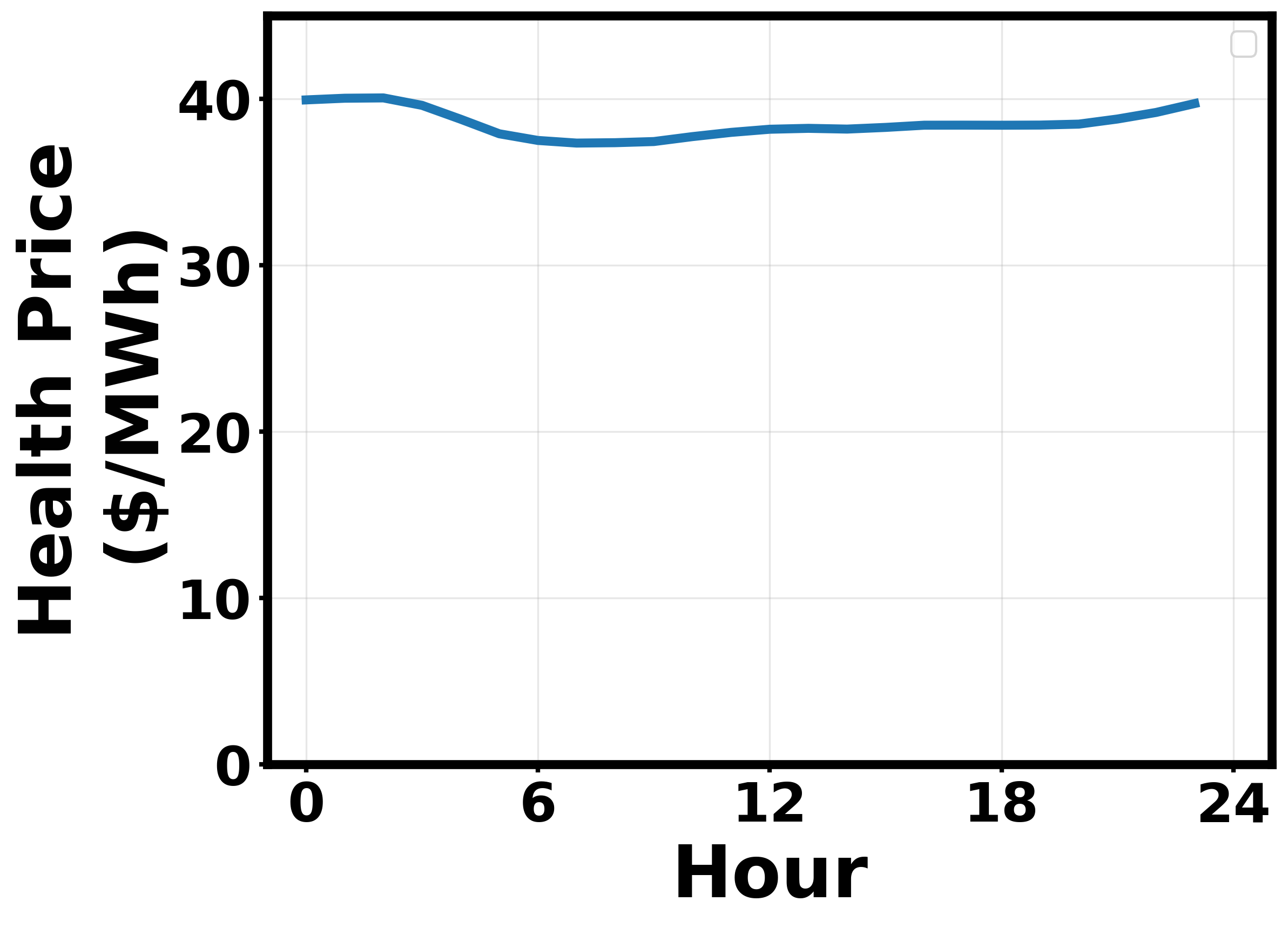}\label{fig:pjm-health-hourly}
        }
    }
    \subfloat[ERCO]{
        {\includegraphics[height=0.21\textwidth]{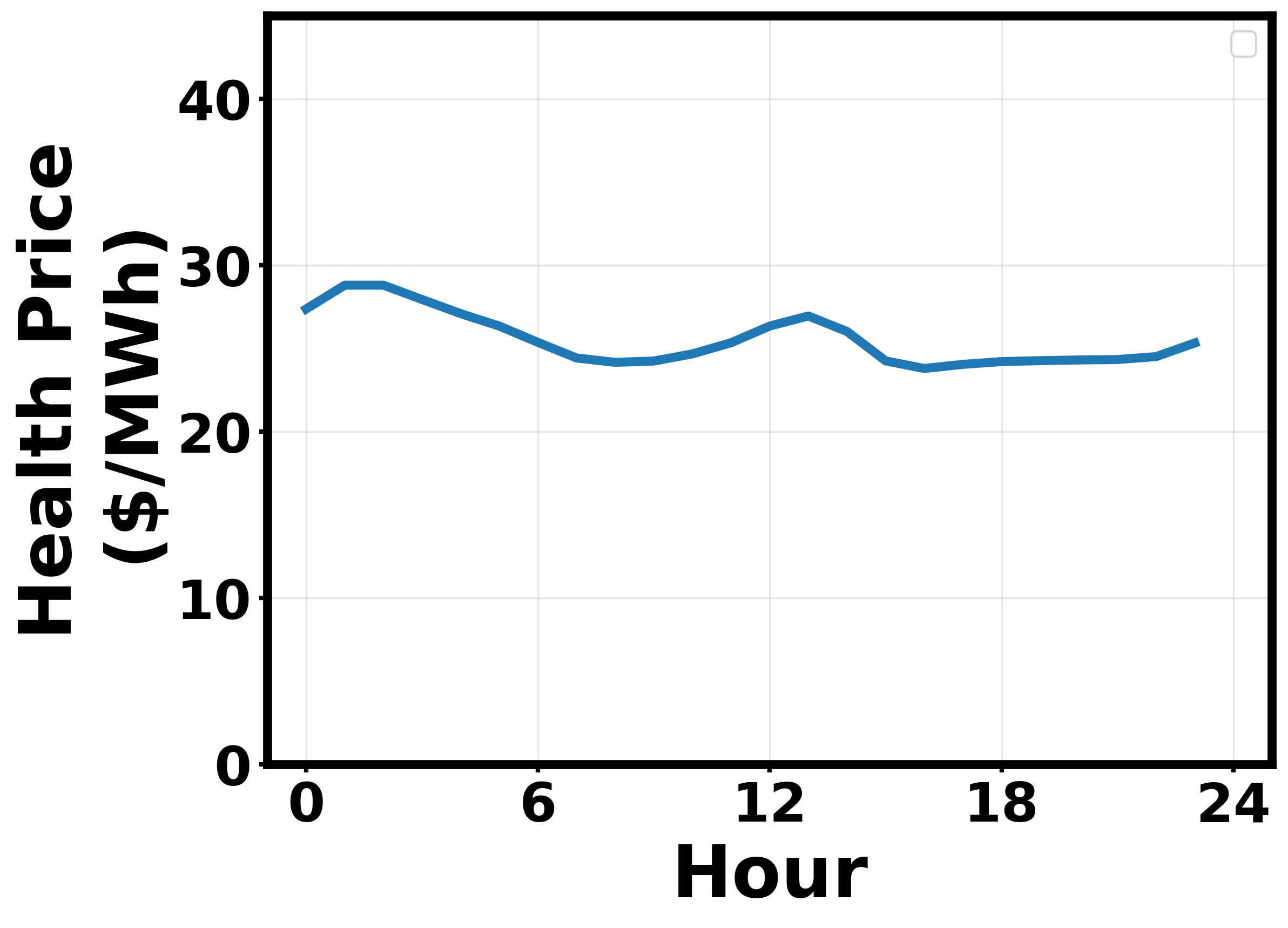}\label{fig:erco-health-hourly}
        }
    }
    \caption{Percentage of hourly health price in CISO, PJM, and ERCO.}
    
    \label{fig:region-health-price-hourly}
\end{figure}

\paragraph{Step II: Derivation of Emission Data}

Emission data were derived from the Environmental Protection Agency’s Emissions \& Generation Resource Integrated Database (eGRID)~\cite{EPA-eGRID}, which provides raw plant-level electricity generation and emission data. Our analysis focuses on four criteria air pollutants: PM\textsubscript{2.5}, SO\textsubscript{2}, NO\textsubscript{X}, and VOC. We obtain plant-level emission data from the most recent available eGRID records: PM\textsubscript{2.5} and VOC from 2021, and SO\textsubscript{2} and NO\textsubscript{X} from 2022. Since eGRID associates each generation facility with its corresponding balancing authority (BA), we construct BA-level emission rates by filtering and aggregating the plant-level data for the three BAs in our study: CISO (California), ERCO (Texas), and PJM (Mid-Atlantic). To validate our derived BA-level rates, we compare them with standard estimates in Table~\ref{tab:emission_rates}. As a baseline verification, we first calculated emission rates by eGRID subregion; these values match the official eGRID Subregion figures exactly, confirming the accuracy of our aggregation methodology. Consequently, our primary calculations by balancing authority (BA) are broadly consistent with the available eGRID data. The observed deviations from AVERT data are expected, as AVERT reports 2023 marginal rates (reflecting the response to load changes) for aggregated balancing authority regions~\cite{EPA-AVERT}, whereas our analysis utilizes 2021/2022 average rates specific to Balancing Authority boundaries.

\begin{table}[ht]
\centering
\caption{Comparison of Emission Rates (lbs/MWh). The AVERT Region column denotes marginal emission rates for aggregated balancing authority regions. The eGRID Subregion column presents annual average output rates at the subregion level, based on the most recent official data (2022 for NO\textsubscript{x}/SO\textsubscript{2}; 2021 for PM\textsubscript{2.5}/VOC). Our calculations are performed at the balancing authority level; for validation, when aggregated by subregion, our results perfectly match the official eGRID figures.}
\label{tab:emission_rates}
\begin{tabular}{l|l|c|c|c|c}
\toprule
Region & Pollutant & \begin{tabular}[c]{@{}c@{}}AVERT\\Region\end{tabular} & \begin{tabular}[c]{@{}c@{}}eGRID\\Subregion\end{tabular} & \begin{tabular}[c]{@{}c@{}}Our Calculation\\(by Subregion)\end{tabular} & \begin{tabular}[c]{@{}c@{}}Our Calculation\\(by Balancing Authority)\end{tabular} \\
\midrule
California & NOx & 0.136 & 0.467 & 0.467 & 0.357 \\
           & SO2 & 0.022 & 0.030 & 0.030 & 0.010 \\
           & PM2.5 & 0.037 & 0.028 & 0.028 & 0.023 \\
           & VOC & 0.014 & 0.019 & 0.019 & 0.012 \\
\midrule
Texas & NOx & 0.624 & 0.468 & 0.468 & 0.432 \\
      & SO2 & 0.565 & 0.407 & 0.407 & 0.337 \\
      & PM2.5 & 0.065 & 0.048 & 0.048 & 0.040 \\
      & VOC & 0.021 & 0.012 & 0.012 & 0.011 \\
\midrule
Mid-Atlantic & NOx & 0.467 & N/A & N/A & 0.248 \\
             & SO2 & 0.530 & N/A & N/A & 0.309 \\
             & PM2.5 & 0.078 & N/A & N/A & 0.046 \\
             & VOC & 0.019 & N/A & N/A & 0.009 \\
\bottomrule
\end{tabular}
\end{table}

To verify the plausibility of our derived health prices, we calculated the health impacts for a hypothetical 10 GWh load using the COBRA model. Our calculations yield average health costs of \$260,000 for CISO and ERCO, and \$390,000 for PJM. For comparison, AVERT’s marginal-based estimates are \$100,000 (California), \$400,000 (Texas), and \$700,000 (Mid-Atlantic). These deviations are consistent with the underlying differences in emission rate definitions, confirming the reliability of our health impact pipeline.

For the selected BA, we then assume a unit electricity consumption (1 MWh) for each hour throughout 2023. Using hourly generation fuel mix data, we allocate this unit hourly demand across different fuel sources according to their generation shares. For each fuel type, we further distribute its allocated generation among all plants within the BA based on their relative generation capacities.

One challenge in this process arises from the inconsistent categorization of fuel types between the EIA and eGRID datasets. To address this, we develop a systematic mapping approach. First, we utilize eGRID's internal hierarchical classification system to map its numerous detailed fuel types to a smaller set of fuel type categories defined within eGRID. Then, we map these simplified eGRID categories to EIA's classification system through both direct correspondence (e.g., HYDRO to WAT) and careful examination of category definitions for less straightforward cases (e.g., DFO to OIL). This meticulous mapping process is essential to ensure accurate integration of emissions data with generation profiles. By combining these carefully harmonized plant-level allocations with plant-specific emission factors, we quantify the specific emissions profile per MWh for each hour.

\paragraph{Step III: Health Cost Assessment}
To assess the public health implications of our emissions profile, we employ the CO-Benefits Risk Assessment (COBRA) Health Impacts Screening and Mapping Tool (Desktop v5.1, as of October 2024) developed by the U.S. EPA~\cite{EPA-COBRA}. COBRA utilizes a reduced-complexity air quality dispersion model incorporating a source-receptor matrix for expedited assessment. Despite its wide validation and adoption in the literature for large-scale air quality and health impact analyses~\cite{schmitt2024health, guidi2024environmental}, applying COBRA to derive health costs requires significant effort. It involves labor-intensive steps to compile and prepare the input data, including mapping emissions profiles to specific regions and ensuring the appropriate application of emission factors, all while maintaining the integrity of the tool’s assumptions. In derivation, we set 2023 as the baseline scenario year to correspond to our study period. In accordance with EPA recommendations based on the U.S. Office of Management and Budget Circular No. A-4 guidance~\cite{OMBA4}, we implement a discount rate of 2\% in the COBRA model. 

Considering the air pollutant transport mechanisms, we account for both internal and external health impacts in our analysis of an emission source. The spatial delineation of each BA's service territories is obtained from the U.S. Energy Atlas~\cite{EIA-Atlas}, which provides raw data in GeoJSON format. To efficiently process these complex geographical data, we employ spatial indexing techniques to optimize the computational performance of point-in-polygon operations, enabling the precise identification of counties within each BA's operational domain. Following the spatial categorization of counties as either internal or external to the BA, we aggregate the county-level health costs accordingly. Specifically, we compute the hourly internal and external health costs throughout the year based on the unit hourly electricity consumption (1 MWh), where internal costs represent the sum of health impacts in counties within the BA's jurisdiction, and external costs comprise impacts in all other counties. This yields a comprehensive dataset comprising hourly fuel mix compositions and their corresponding internal and external health costs per MWh for the entire year, which is used to train our proposed pipeline.

\paragraph{Brief Summary of Dataset Preparation Challenges}  Challenges in dataset preparation mainly include reconciling semantic inconsistencies across data sources, resolving spatial mismatches, and managing infrastructure for local health impact computation. For example, the U.S. EIA defines 8 fuel mix categories, while the EPA’s eGRID lists over 40, requiring us to systematically consolidate and map these types using internal hierarchies and cross-referencing definitions. Spatial integration is equally nontrivial—health impacts from COBRA are county-based, whereas eGRID data is organized by BAs. To bridge this gap, we employ U.S. Energy Atlas GeoJSON files with optimized spatial indexing for accurate point-in-polygon assignments. Additionally, estimating county-level health costs using COBRA's desktop tool demanded significant manual effort, with each run taking 5–20 minutes and data entry requiring 1–2 minutes per input. This entire process spans several days and underscores the  effort involved in building a reliable, multi-source dataset.

\subsection{Additional Empirical Details and Results}\label{appendix:additional_empirical} 
The three regions—CISO (California), PJM (Mid-Atlantic), and ERCO (Texas) selected in our main text have shown the effectiveness of our methods in various energy generation patterns and regulatory environments. These regions represent distinct characteristics in power grid operations, emissions profiles, and public health impact patterns. Texas, for example, ranks among the top three for PM\textsubscript{2.5} emissions, which have severe health effects~\cite{EPA_AirPollution_CrossState_Website}. Although California does not have the highest emissions, its dense population results in significant adverse health costs~\cite{epa2019bpk}. Specifically, the EPA reports relatively low health benefits per kWh from energy efficiency and renewable energy in California~\cite{epa2019bpk}.

\textbf{Model Construction}\quad
For the fuel-mix predictor, we develop a Transformer-based architecture tailored to fuel mix time-series data, capitalizing on its ability to capture intricate relationships between various factors influencing the fuel mix. The transformer also excels at capturing long-term dependencies, which are critical in understanding the temporal dynamics of fuel usage and transitions over extended periods. The architecture consists of an embedding layer followed by a Transformer block with a single encoder and decoder layer, utilizing four multi-head attention mechanisms with a dropout regularization rate of 0.1.

The conversion of pollutant emissions to air pollutant concentrations and their subsequent dispersion in the atmosphere is a highly intricate process. It involves complex chemical transformations, atmospheric reactions, and meteorological processes. To address this complexity, we utilize a 3-layer Multi-Layer Perceptron (MLP) model, which takes the fuel mix predictions as input and predicts the potential health impact.  The model is specifically chosen for its ability to approximate complex, nonlinear relationships inherent in pollutant dispersion and their effects. 

In experiments, the LSTM based fuel mix predictor is composed of an embedding layer that projects inputs to a 64-dimensional space, followed by a single-layer LSTM with 64 hidden units and a dropout rate of 0.1. The number of training epochs is set to 100 for Transformer-based methods, while it is set to 100 for the LSTM architecture.

\textbf{Implementation Details}\quad In our experiments, predictions are made across different time window steps, denoted as $T$. We set \( T \) to values of 24 and 72 hours to explore the impact of varying prediction time windows. Temporal sequences are handled by slicing inputs and targets according to these specified sliding window steps $T$. We employ an 80/20 train-test split, where a portion 
of the training set is reserved for validation to tune hyperparameters, 
while the test set remains strictly held-out. For the CISO region dataset, we utilize the Stochastic Gradient Descent (SGD) optimizer with a learning rate of 0.004 and a batch size of 128. In addition to our Transformer-based models, we also implement LSTM-based variants of both the Fuel-mix-driven Opt and Health-driven Opt methods as \textbf{baselines} for comparison. These LSTM baselines use the same optimization objectives as their corresponding Transformer-based counterparts, i.e., Health-driven Opt and Fuel-mix-driven Opt, respectively. All experiments are conducted on a single NVIDIA K80 GPU. Training the Transformer-based models for 100 epochs takes usually one hour.

\end{document}